\documentclass[letterpaper]{article} 
\usepackage{aaai2026} 
\usepackage{times}  
\usepackage{helvet}  
\usepackage{courier}  
\usepackage[hyphens]{url}  
\usepackage{graphicx} 
\usepackage{natbib}  
\usepackage{caption} 
\usepackage{amsmath} 
\usepackage{multirow}
\usepackage{booktabs} 
\usepackage{dblfloatfix}
\usepackage{mathtools}
\usepackage[table]{xcolor}
\usepackage[dvipsnames]{xcolor}
\usepackage{tabularx} 
\usepackage{xspace}
\usepackage{amssymb}

\usepackage{subcaption}
\usepackage{float}
\usepackage{multicol}
\usepackage{pifont}  

\usepackage{algorithm}
\usepackage{algorithmic}
\usepackage{xcolor}
\usepackage{soul}

\usepackage{newfloat}
\usepackage{listings}
\DeclareCaptionStyle{ruled}{labelfont=normalfont,labelsep=colon,strut=off} 
\floatstyle{ruled}
\newfloat{listing}{tb}{lst}{}
\floatname{listing}{Listing}
\title{LOGicalThought: Logic-Based Ontological Grounding of LLMs for High-Assurance Reasoning}

\author{
  Navapat Nananukul\textsuperscript{\rm 1},
  Yue Zhang\textsuperscript{\rm 2},
  Ryan Lee\textsuperscript{\rm 1},
  Eric Boxer\textsuperscript{\rm 1},
  Jonathan May\textsuperscript{\rm 1},
  Vibhav Giridhar Gogate\textsuperscript{\rm 2},
  Jay Pujara\textsuperscript{\rm 1},
  Mayank Kejriwal\textsuperscript{\rm 1}
}
\affiliations{
  \textsuperscript{\rm 1} University of Southern California\\
  \textsuperscript{\rm 2} The University of Texas at Dallas
}

\begin{document}

\maketitle

\begin{abstract}

High-assurance reasoning, particularly in critical domains such as law and medicine, requires conclusions that are accurate, verifiable, and explicitly grounded in evidence. This reasoning relies on premises codified from rules, statutes, and contracts, inherently involving defeasible or non-monotonic logic due to numerous exceptions, where the introduction of a single fact can invalidate general rules, posing significant challenges. While large language models (LLMs) excel at processing natural language, their capabilities in standard inference tasks do not translate to the rigorous reasoning required over high-assurance text guidelines. Core reasoning challenges within such texts often manifest specific logical structures involving negation, implication, and, most critically, defeasible rules and exceptions. In this paper, we propose a novel neurosymbolically-grounded architecture called LOGicalThought (LogT) 
that uses an advanced logical language and reasoner in conjunction with an LLM to construct a dual \textit{symbolic graph context} and \textit{logic-based context}. These two context representations transform the problem from  inference over long-form guidelines into a compact grounded evaluation. Evaluated on four multi-domain benchmarks against four baselines, LogT improves overall performance by 11.84\% across all LLMs. Performance improves significantly across all three modes of reasoning: by up to +10.2\% on negation, +13.2\% on implication, and +5.5\% on defeasible reasoning compared to the strongest baseline.

\end{abstract}

\section{Introduction}

In recent years, the deployment of large language models (LLMs) for tasks demanding complex reasoning has marked a significant advancement in prompting approaches like Thought Generation, Program-Aid, and Self-Criticism \cite{DBLP:conf/nips/Wei0SBIXCLZ22,pmlr-v202-gao23f,DBLP:conf/acl/DhuliawalaKXRLC24}. However, LLMs have seen limited evaluation in \emph{high-assurance domains} such as tax law and healthcare, where reasoning must be rigorous, verifiable, and explicitly evidence-based. \cite{groen2014relationship,susskind1986expert}. 
While methods like Chain-of-Thought (CoT) prompting can generate reasoning chains in an end-to-end fashion, they lack the rigor of strict logic symbolic derivation, leading to potentially inaccurate conclusions \cite{lyu2023faithful,zhang2024chain,zhou-etal-2021-rica}. Also, their reasoning chains are not grounded as formally verifiable claims, making it difficult to assess whether their reasoning was, in fact, correct, even when the answer itself is correct \cite{turpin2023language,lanham2023measuring}. 

In this work, we introduce a neurosymbolic framework called LOGicalThought (\textsc{LogT}) aimed at significantly enhancing the high-assurance reasoning capabilities of LLMs.
\textsc{LogT} 
addresses current challenges with high-assurance reasoning by grounding the original natural language problem into a \textit{dual} neuro-symbolic context extracted from long-form guidelines: an ontologically-grounded \textit{symbolic graph} capturing the high-level relationships in the text, and a machine-readable \textit{logic program} formalizing the underlying defeasible rules and facts. By providing the dual context to the LLM, \textsc{LogT} enables a more robust and transparent reasoning process. Specific contributions include:
\begin{itemize}

\item A formal specification for high-assurance reasoning that decomposes and extends the (originally, natural language) inference task via a \textit{logic program mapping} that explicitly represents non-monotonic logical constructs.

\item A novel neurosymbolically guided approach (\textsc{LogT}) that addresses the challenges of high-assurance reasoning by constructing a \textit{dual context} based on a symbolic graph representation and a non-monotonic logic program. 


\item A systematic benchmark construction methodology for augmenting ordinary inference datasets with scenarios that enable LLM testing on negation, implication, and defeasibility. Additionally, we introduce a new, hand-crafted benchmark based on a complex text game. 


\item Extensive evaluation of \textsc{LogT}'s accuracy and reasoning performance on four expert-level benchmarks against four competitive LLM-based baselines. \textsc{LogT} is found to produce higher-assurance reasoning traces compared to CoT, and is also more accurate.

\end{itemize}

\textsc{LogT} follows a long tradition in AI of building systems capable of logical deduction that extends significantly beyond ordinary propositional and first-order logic (e.g., through defeasible and non-monotonic reasoning).  While much recent success has been achieved with LLMs in informal ``common sense'' reasoning settings, and problems with clear reasoning pathways (such as high school mathematics), non-monotonic and defeasible reasoning has been less studied (see \textit{Related Work}). We aim to address this gap through both our proposed neurosymbolic approach and evaluation methodology.

\section{Problem Formulation}


Given a set of guidelines $X$ and an associated scenario $S$ providing additional context, the task is to evaluate the truth of a hypothesis $H$ by reasoning from $X$ and the facts stated in $S$. An example of all three can be found in the left column of Figure 1, based on a high-assurance task (tax law). 

In this paper, we consider the following three (non-exhaustive) modalities of hypotheses for evaluating high-assurance reasoning: \textit{negation, implication,} and \textit{defeasible}. We formulate our targeted problem as predicting the logical relation $\hat{L}= \operatorname{M}( f(X, S, H))$ between $H$ and $(X, S)$. 


Here, $\hat{L} \in \{\texttt{Entailment},\texttt{Contradiction},\texttt{Neutral}\}$ indicates whether $H$ is entailed, contradicted, or neutral given $X$ and $S$. The function $f$ extracts logic program rules from the natural language inputs $(X, S, H)$, decomposing the problem into three reasoning modes:
\begingroup\small
\begin{align*}
    f_{\mathrm{neg}}(X, S, H) =\;\;& 
      \big\{\, r_i\!:~ \neg\,\mathrm{q}_i \leftarrow \mathrm{p}_i \,\big\} \\[0.5em]
   f_{\mathrm{imp}}(X, S, H) =\;\;& 
      \big\{\, r_i\!:~ \mathrm{q}_i \leftarrow \mathrm{p}_i\,\big\} \\[0.5em]
    f_{\mathrm{def}}(X, S, H) =\;\;& 
      \big\{\, r_i\!:~ \mathrm{q}_i \leftarrow \mathrm{p}_i, \\
    &\hspace{0.7em} r_j\!:~ \neg\,\mathrm{q}_i \leftarrow \mathrm{exception}_j, \\
    &\hspace{0.7em} \text{exception overrides default $p_i$} \,\big\}
\end{align*}
\endgroup

Here, $\mathrm{p}_i$ is the condition for a rule, $\mathrm{q}_i$ is the conclusion, both extracted from $(X, S, H)$, and $\mathrm{exception}_j$ is a condition for defeasible reasoning when a natural language exception overrides a default condition. In Appendix D, we provide a table with illustrative examples of all three modes.


In an evaluation setting, a model $M$ is considered successful if its prediction $\hat{L}$ matches the truth-logical relationship $L$. Note that $M$ and $f$ do not necessarily have to be LLM-based, and also do not need to be explicitly specified. For example, ordinary LLM prompting implicitly models $f$, whereas a more neurosymbolic approach, such as our proposed approach, would aim to make $f$ more explicit. The presence of logic rules $r_i, r_j \in R$ within a model's generated reasoning trace is considered direct evidence that its conclusion is grounded by the logic program generated from function $f$.

\begin{figure*}[t]
  \centering
  \includegraphics[width=0.8\textwidth]{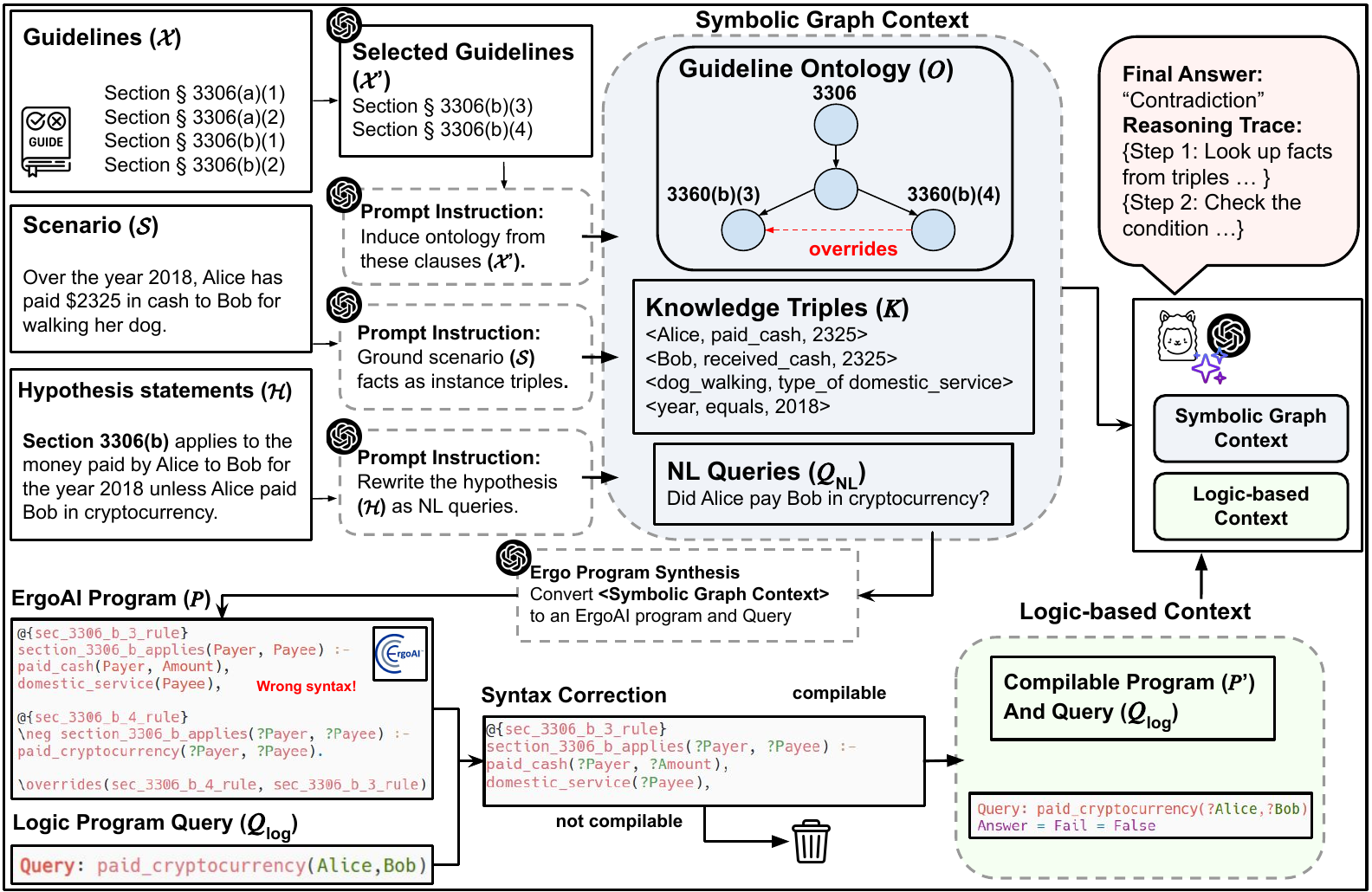}
  \caption{A schematized illustration of \textsc{LogT}. An LLM initially processes raw guidelines, a scenario, and a hypothesis to create a \textit{symbolic graph context}. This unified structure contains a rule ontology, factual knowledge triples, and a natural language query. This context acts as an input for two distinct reasoning approaches: (1) it provides rich context for guiding LLM reasoning, and (2) it serves as a blueprint for synthesizing a logic program. Compilable logic programs and queries are formalized into \textit{logic-based context}, completing the dual neurosymbolic context underlying the approach.}
  \label{fig:logt}
\end{figure*}


\section{LogicalThought}


\label{method}

\begin{algorithm}[t]
\caption{The \textsc{LogT} Grounded Reasoning Framework}
\label{alg:logt}
\small
\begin{algorithmic}[1]
\REQUIRE LLM; Guidelines $X$; Scenario $S$; Hypothesis $H$
\ENSURE Predicted Label $\hat{L}$; Reasoning Trace $T$

 \textcolor{Gray!80!black}{\textit{--- Stage 1: Symbolic Context Generation ---}}
\STATE $X'_{\text{}} \gets \text{RulesSelection}(X, S, H)$
\STATE \hfill \COMMENT{\textcolor{olive}{Filter rules from $X$ relevant to $S, H$}}
\STATE $\mathcal{O}_{\text{rules}}, \mathcal{K}_{\text{instance}}, \mathcal{Q}_{\text{nl}} \gets \text{SymbolicContext}(X'_{\text{}}, S, H)$
\STATE \hfill \COMMENT{\textcolor{olive}{Generate Ontology, Triples, and NL Queries}}
\STATE $\mathcal{C}_{\text{sym}} \gets \{ \mathcal{O}_{\text{rules}}, \mathcal{K}_{\text{instance}}, \mathcal{Q}_{\text{nl}} \}$

 \textcolor{Gray!80!black}{\textit{--- Stage 2: Logical Context Construction ---}}
\STATE $\mathcal{F} \gets \text{SynthesizeFacts}(\mathcal{K}_{\text{instance}})$
\STATE $\mathcal{R}_{\text{base}}, \mathcal{R}_{\text{override}} \gets \text{SynthesizeRules}(\mathcal{O}_{\text{rules}})$
\STATE $\mathcal{Q}_{\text{Ergo}} \gets \text{SynthesizeQueries}(\mathcal{Q}_{\text{nl}})$
\STATE \hfill \COMMENT{\textcolor{olive}{All synthesis via LLM with ErgoAI syntax template}}
\STATE $\mathcal{P} \gets \mathcal{F} \cup \mathcal{R}_{\text{base}} \cup \mathcal{R}_{\text{override}}$ \STATE \hfill \COMMENT{\textcolor{olive}{Assemble the full program}}
\STATE $\mathcal{P} \gets \text{SyntacticCorrection}(\mathcal{P})$ \STATE \hfill \COMMENT{\textcolor{olive}{Apply rule-based script}}
\STATE $\mathcal{P'} \gets \text{ErgoAI\_CompileAndFilter}(\mathcal{P})$ \STATE \hfill \COMMENT{\textcolor{olive}{Keep only compilable part of the program}}
\STATE $\mathcal{A} \gets \text{ErgoAI\_ExecuteQueries}(\mathcal{P'}, \mathcal{Q}_{\text{Ergo}})$
\STATE $\mathcal{C}_{\text{log}} \gets \{ \mathcal{P'}, \mathcal{A} \}$

 \textcolor{Gray!80!black}{\textit{--- Stage 3: Hypothesis Evaluation ---}}
\STATE $(\hat{L}, T^{\text{raw}}) \gets \mathcal{M}(\mathcal{C}_{\text{sym}}, \mathcal{C}_{\text{log}}, H)$
\STATE \hfill \COMMENT{\textcolor{olive}{Grounded LLM call, see prompt in Appendix C.5}}
\STATE $T_i \gets \text{OrganizeTrace}(T^{\text{raw}};\, \mathcal{T}_{\text{templates}})$
\STATE \hfill \COMMENT{\textcolor{olive}{Re-organize reasoning into six trace types according to a pre-defined template}}
\STATE \textbf{return} $(\hat{L}, T)$
\end{algorithmic}
\end{algorithm}





The central idea of \textsc{LogT} is to convert complex, unstructured reasoning tasks into a concise, ontologically grounded neurosymbolic representation, facilitating robust inference. Unlike much of the prior work (such as in the growing literature on graph-based retrieval augmented generation \cite{peng2024graph,li2024simple}), where the neurosymbolic context is typically some version of a knowledge graph, the neurosymbolic context extracted by \textsc{LogT} from raw natural language inputs comprises both knowledge \textit{and} logic. This transformation is essential in high-assurance domains, where reasoning must navigate the subtleties, exceptions, and complex dependencies codified in their guidelines. Such challenges often manifest as specific logical structures involving negation, implication, and defeasible reasoning \cite{xiu2022logicnmr}. LLMs can be brittle in such non-monotonic settings, with their performance notably deteriorating when required to reason through extensive and intricate rule sets. \cite{li2023long, li2023loogle}.

\paragraph{Symbolic graph context.}
To better guide LLMs, \textsc{LogT} first constructs a \textit{Symbolic Graph Context} ($\mathcal{C}_{\text{sym}}$) in a two-step process. First, to overcome the challenge of long guidelines, an LLM is prompted to select only the pieces of information needed to resolve the given scenario and hypothesis. Formally, given the raw input consisting of the guideline $X$, scenario $S$, and hypothesis $H$, an LLM first performs \textit{guideline selection} to select only the subset of rules $X'$ from $X$ that are relevant for resolving $H$ given $S$. 

Next, a prompting step performs three related graph-extraction tasks simultaneously: 
\begin{enumerate}
\item \textbf{Guideline Ontology ($O_{rules}$)}: A graph-based representation of the selected rules from the guidelines-set $X$, capturing concepts, relations and hierarchical dependencies encoded in the selected guidelines.
\item \textbf{Knowledge Triples ($K_{instance}$)}: A set of structured triples of the form (subject, predicate, object) encoding entities and relations in the scenario $S$ and hypothesis $H$, constrained and typed according to the ontology $O_{rules}$.
\item \textbf{Queries ($Q_{nl}$)}: The original hypothesis $H$ decomposed into a set of precise, natural language questions that must be resolved to evaluate the hypothesis for each reasoning mode.
\end{enumerate}


Together, these three components form the complete \textit{symbolic graph context} for the instance: $$\mathcal{C}_{\text{sym}} = \{ \mathcal{O}_{\text{rules}}, \mathcal{K}_{\text{instance}}, \mathcal{Q}_{\text{nl}} \}$$

\paragraph{Logic-based context.} To capture the exceptions required for non‑monotonic reasoning, \textsc{LogT} also synthesizes a formal \textit{logic-based context}. An LLM translates the symbolic graph context $\mathcal{C}_{\text{sym}}$ into a logic program: facts are derived from knowledge fragments $K_{instance}$, rules and (defeasible) overriding rules are generated from the ontology $O_{rules}$, and the natural‑language questions $Q_{nl}$ are converted into executable queries. Ideally, the objective of this stage is to produce a \emph{machine‑readable} knowledge base (KB) that stores facts and rules and correctly returns results when queried. We chose ErgoAI, based on RuleLog \cite{grosof2017rulelog}, for our logic and reasoning engine because it is designed for higher-order logic and non-monotonic reasoning with built-in features that extend far beyond standard logic programming \cite{grosof2023ergo}. ErgoAI's native support for defeasible reasoning allows it to directly execute the kind of exception-driven logic that is fundamental to our high-assurance domains.

We utilize ErgoAI’s built‑in functionality for facts, rules, and defeasible reasoning with exception handling to validate the KB. The resulting ErgoAI program then undergoes a two‑stage verification: (i) a rule‑based script applies syntactic corrections; and (ii) the corrected program is compiled and executed in the ErgoAI reasoner. We retain only the compilable subset and execute the queries to obtain zero, one, or multiple answers to the queries. Formally, as also shown in Algorithm 1, the LLM maps $(O_{rules}, K_{instance}, Q_{nl})$ into an ErgoAI logic program $P$ using a few‑shot ErgoAI syntax rule bank (facts, rules, defeasible rules; illustrative examples are provided in Appendix E).  The LLM is then instructed to perform the following mappings:

\begin{enumerate}
\item \textbf{Facts ($F$)}: Convert $K_{instance}$ into atomic predicates; entity identifiers are canonicalized and constants are typed to match the predicate schema.
\item \textbf{Rules and Defeasible Rules ($R_{base}, R_{override}$)}: Translate hierarchical relationships from $O_{rules}$ into rules and defeasible (overriding) rules in ErgoAI format, encoding priorities from the ontology’s partial order.
\item \textbf{Logic Program Queries ($Q_{log}$)}: Convert $Q_{nl}$ into formal logic queries with explicit variables.
\end{enumerate}
Because the LLM‑generated program $P$ may contain errors, we apply automated syntactic fixes (e.g., terminators, variable normalization) and then compile the program in the ErgoAI reasoner. We construct the compilable program $P'$ by retaining only facts and rules that successfully compiled and discard the rest. We then execute $Q_{Ergo}$ against $P'$ and store any resulting answers $A$. The final \textit{logic-based context} is $$ C_{log} = \{ P', A \}.$$

If query compilation fails, we fall back to using only the verified program, $\mathcal{P'}$, as the context. Finally, both the \textit{symbolic context} ($\mathcal{C}_{\text{sym}}$) and the \textit{logic-based context} ($\mathcal{C}_{\text{log}}$) are provided to ground the LLM for the final prediction. The prompt templates and configurations, including fixed seeds and temperature for reproducibility, are detailed in Appendix B and C.3. Because discarding non-compilable parts of the original program $P$ could potentially remove important information, we include within the final context the entire verified program, $P'$ rather than relying solely on the query answers $A$ produced by $P'$.

\paragraph{Hypothesis evaluation using neurosymbolic context}

The final stage of \textsc{LogT} involves using the two contexts constructed in the previous steps to ground LLMs for evaluating hypothesis $H$. We define $\mathcal{M}$ as the LLM being evaluated on the entailment task. $\mathcal{M}$ accepts the two contexts ($\mathcal{C}_{\text{sym}}$ and  $\mathcal{C}_{\text{log}}$) and the original hypothesis as an input and is prompted to produce its predicted label (from one of \textit{entailment, contradiction,} and \textit{neutral}), and a reasoning trace: $$\hat{L}, T^{raw} = \mathcal{M}(C_{sym}, C_{log},  H)$$ 

By grounding the LLM using both contexts, we hypothesize that it will enable the LLM to conduct a focused semantic evaluation of pre-compiled knowledge \textit{vis-a-vis} open-ended reasoning over long-form documents. We provide the definition of \texttt{entailment} if the contexts corresponding to the hypothesis are derivable, \texttt{contradiction} if the negated context is derivable, and \texttt{neutral} otherwise. 

To enable a standard comparison between the reasoning trace $T^{raw}$ output by \textsc{LogT} and that output by baselines such as CoT, we use an LLM-as-a-judge \cite{zheng2023judging,gu2024survey} framework to re-organize each raw generated reasoning step into one of six types: \textit{fact\_lookup}, \textit{apply\_rule}, \textit{check\_condition}, \textit{resolve\_conflict/override}, \textit{contradiction\_detected}, and \textit{conclude\_label}. Subsequently, we analyze how the reasoning trace from \textsc{LogT} is aligned with the symbolic and logical contexts by linking the \textit{fact\_lookup} and \textit{apply\_rule} trace types to their provenance in $O_{rules}$ and $K_{instance}$.

\section{Experimental Setup}

\paragraph{Benchmarks.}
We evaluate \textsc{LogT} on three established NLI benchmarks as well as a new hand-crafted benchmark incorporating domain expert guidelines, scenarios, and hypotheses. These benchmarks pose significant challenges, primarily requiring identification of pertinent evidence and inference over dense, complex textual guidelines. \textbf{ContractNLI} is a legal domain NLI dataset with document-level evidence in contract clauses and three labels: entailment, contradiction, and neutral \cite{koreeda-manning-2021-contractnli-dataset}. \textbf{Statutory Reasoning Assessment (SARA)} offers scenarios for statutory reasoning using the US Internal Revenue Code, requiring the application of complex legal rules \cite{DBLP:conf/kdd/HolzenbergerBD20}. The \textbf{Safe Biomedical NLI for Clinical Trials (BioMedNLI)} evaluates reasoning in medical contexts by classifying related statements as entailment or contradiction based on clinical trial reports \cite{jullien-etal-2024-semeval}. Lastly, \textbf{Dungeons \& Dragons NLI} is a custom benchmark proposed in this paper that is drawn from the Dungeons \& Dragons rulebook, also featuring scenarios, rules, and test hypotheses. Detailed statistics for all benchmarks are provided in Appendix F.   






\paragraph{Benchmark Enhancement for Non-Monotonic Reasoning.}

Existing entailment benchmarks typically contain hypotheses that are relatively simple and are based on associations with prior knowledge acquired during an LLM's pre-training owing to their longstanding status. While some benchmarks include examples involving negation, implication, or defeasible, they usually do so implicitly and in isolation, not as part of a unified categorization. Hence, current NLI benchmarks do not support proper evaluation of how well rival models handle non-monotonic reasoning modes. It remains unclear whether a model can robustly distinguish, for example, between a hypothesis that contradicts the premise due to negation, versus one that is overridden by an exception (defeasibility).
Our preliminary analysis shows that current domain-expert NLI benchmarks are insufficiently challenging for advanced LLMs, suggesting potential exposure during model pre-training or inadequately complex tasks\footnote{In a pilot study, we found newer LLMs to achieve near-perfect accuracy (0.90–0.95) on existing NLI benchmarks just from basic prompting. We provide the preliminary results in Appendix G}. For newer LLMs, likely due to pre-training bias or due to inadequate complexity of the tasks themselves \cite{yuan2023revisiting}. 

We introduce a systematic augmentation pipeline (Figure \ref{fig:bench-gen}) that organizes hypotheses under one of three reasoning modes: \textit{negation, implication}, and \textit{defeasibility}. As alluded to earlier, these modes (while not exhaustive to all high-assurance reasoning) are broadly encountered in human reasoning and decision-making, prove to be challenging for LLMs, and have minimal representation in current NLI benchmarks. For each original hypothesis $hyp_i$, the methodology can generate three hypothesis according to the following operative definitions:

\begin{figure}[t]
    \centering
    \includegraphics[width=\columnwidth]{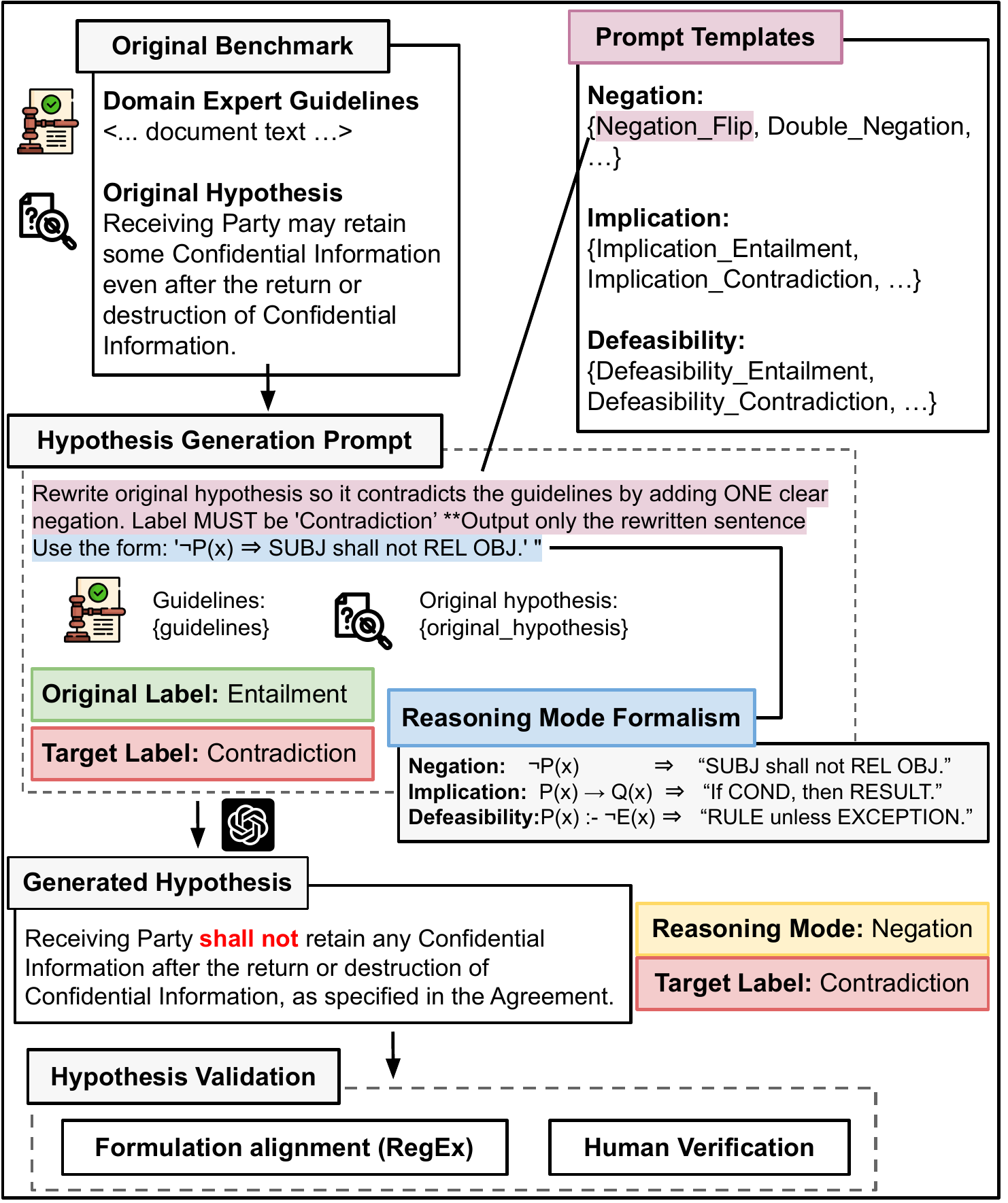}
    \caption{The proposed benchmark enhancement workflow for evaluating three modes of reasoning (\textit{negation, implication}, and \textit{defeasibility}).}
    \label{fig:bench-gen}
\end{figure}

\begin{enumerate}
  \item \textbf{Negation} ($h_i^{\mathrm{neg}}$)\, is a claim that directly contradicts a rule-derived fact, typically expressed in the form “[Subject] shall not [Relate to] [Object],” requiring the system to identify when a prohibition overrides an allowance.
  \item \textbf{Implication} ($h_i^{\mathrm{imp}}$)\, is a conditional claim that asserts a cause-effect relationship based on the rule, expressed as “If [Condition], then [Result],” requiring the system to reason over hypothetical or contingent outcomes.
  \item \textbf{Defeasibility} ($h_i^{\mathrm{deaf}}$)\, is a rule-based statement that includes an exception, typically in the form “[Rule] unless [Exception],” requiring the system to understand when a general rule may not apply due to exceptions.
\end{enumerate}

We instantiate these transformations with eleven prompt templates, controlled by three symbolic reasoning modes and their corresponding target label. For example, as illustrated in Figure \ref{fig:bench-gen} , a negation template transforms an original hypothesis labeled as \textit{Entailment} into one classified as \textit{Contradiction}. In contrast, another template applying a double negation increases linguistic complexity while preserving the original label. 

Once generated, the hypotheses are verified using two steps. First, we use a heuristic regular expression function with a pre-defined list of keywords to ensure the intended reasoning modes are present. Second, we confirm that the generated hypotheses are more difficult than the original by running them through two pre-trained NLI models. We provide all results and prompt templates in Appendix C and G, together with a detailed exposition on the verification. Finally, to ensure the quality and correctness of this automated process, we conducted a manual audit on a random sample of 100 generated hypotheses per benchmark, verifying that the new hypotheses are consistently well-formed, and correctly aligned with the semantics and ground-truth labels of the targeted reasoning modes.

\paragraph{Baselines.}

We evaluate \textsc{LogT}'s accuracy against five baselines including: \textit{(1) Basic Prompting with No Document (basic-nd)}, which relies on the LLM to answer the question directly using its own knowledge; \textit{(2) Basic Prompting with Document (basic-d)}, which provides the model with the relevant domain-expert guidelines as added context; \textit{(3) Few-Shot Prompting (FS)} \cite{brown2020language}, where the prompt is constructed with a solved example from a training set for each possible class label (e.g., Entailment, Contradiction) to guide the model via in-context learning ; and (4) \textit{Multi-Step Chain-of-Thought (CoT)} \cite{wei2022chain}, an advanced technique where the model first generates a step-by-step rationale that is then programmatically used in subsequent prompts to synthesize the final answer. 

Additionally, we evaluate the quality of \textsc{LogT}'s reasoning traces in comparison to multi-step CoT reasoning by considering cases where the predicted labels contradict those produced by an alternative method. We use the \textit{LLM-as-a-judge} \cite{zheng2023judging,gu2024survey} framework to categorize each generated reasoning step into one of six types: \textit{fact\_lookup}, \textit{apply\_rule}, \textit{check\_condition}, \textit{contradiction\_detected}, and \textit{conclude\_label}. We provide all prompt template examples in appendix C.1. Each full reasoning trace is then classified as a correct \textit{Reason} if it contains appropriate and coherent use of rules and facts that support the final decision, or as an incorrect \textit{Reason} if the reasoning is flawed, incomplete, or inconsistent.

We use six diverse LLMs including Mistral 7B, LLaMA 3.1-8B, LLaMA 3.1-70B, Claude 3.5 Haiku, GPT-o3 Mini, and DeepSeek R1. A comprehensive table with model details, including their number of parameters, release dates, and developers, is available in Appendix A.
To ensure a fair comparison, all methods (\textsc{LogT} and baselines) are always provided with the same system prompts and in-context examples where applicable. For reproducible and deterministic results, we set the decoding temperature to 0. Performance is evaluated across all three reasoning categories (\textit{implication, negation,} and \textit{defeasibility}).

\begin{table*}[t]
  \centering
  \small
  \setlength{\tabcolsep}{8pt}
  \renewcommand{\arraystretch}{1}
  \begin{tabular}{lccccccc} 
    \toprule
    \textbf{Model}
      & \textbf{Basic-ND}
      & \textbf{Basic-D}
      & \textbf{FS}
      & \textbf{CoT}
      & \textbf{LogT (SGC)}
      & \textbf{LogT (LC)}
      & \textbf{LogT (Full)} \\    
    \midrule

    \rowcolor{gray!15}\multicolumn{8}{c}{\textbf{ContractNLI*}} \\ \midrule
    Mistral 0.3-7B
      & 34.00 {\scriptsize(±1.48)}
      & 44.30 {\scriptsize(±1.58)}
      & 45.20 {\scriptsize(±1.58)}
      & 44.40 {\scriptsize(±1.58)}
      & 47.90 {\scriptsize(±1.59)}  
      & 52.90 {\scriptsize(±1.56)}  
      & \textbf{53.40} {\scriptsize(±1.58)} \\

    LLaMA 3.1-8B
      & 38.00 {\scriptsize(±1.53)}
      & 39.00 {\scriptsize(±1.53)}
      & 40.60 {\scriptsize(±1.53)}
      & 44.30 {\scriptsize(±1.58)}
      & 43.00 {\scriptsize(±1.55)}
      & 47.40 {\scriptsize(±1.55)}
      & \textbf{50.40} {\scriptsize(±1.58)} \\

    LLaMA 3.3-70B
      & 48.60 {\scriptsize(±1.58)}
      & 62.70 {\scriptsize(±1.53)}
      & 64.30 {\scriptsize(±1.53)}
      & 62.70 {\scriptsize(±1.53)}
      & 59.00 {\scriptsize(±1.50)}
      & 62.90 {\scriptsize(±1.53)}
      & \textbf{66.90} {\scriptsize(±1.48)} \\

    Claude 3.5 Haiku
      & 32.40 {\scriptsize(±1.48)}
      & 62.60 {\scriptsize(±1.53)}
      & 61.70 {\scriptsize(±1.53)}
      & 63.90 {\scriptsize(±1.53)}
      & 64.50 {\scriptsize(±1.50)}
      & 65.80 {\scriptsize(±1.45)}
      & \textbf{65.90} {\scriptsize(±1.48)} \\

    GPT-o3 Mini
      & 42.70 {\scriptsize(±1.58)}
      & 69.90 {\scriptsize(±1.43)}
      & \textbf{70.00} {\scriptsize(±1.43)}
      & 61.60 {\scriptsize(±1.53)}
      & 64.30 {\scriptsize(±1.50)}
      & 64.50 {\scriptsize(±1.50)}
      & 69.20 {\scriptsize(±1.48)} \\

    DeepSeek R1
      & 54.40 {\scriptsize(±1.58)}
      & 70.10 {\scriptsize(±1.43)}
      & 69.90 {\scriptsize(±1.43)}
      & 61.30 {\scriptsize(±1.53)}
      & 66.60 {\scriptsize(±1.45)}
      & 68.70 {\scriptsize(±1.44)}
      & \textbf{70.20} {\scriptsize(±1.43)} \\

    \midrule
    \rowcolor{gray!15}\multicolumn{8}{c}{\textbf{SARA*}} \\ \midrule
    Mistral 0.3-7B
      & 65.13 {\scriptsize(±2.94)}
      & 62.84 {\scriptsize(±2.99)}
      & 64.75 {\scriptsize(±2.96)}
      & 63.22 {\scriptsize(±2.99)}
      & 62.80 {\scriptsize(±3.00)}
      & 65.51 {\scriptsize(±2.95)}
      & \textbf{72.41} {\scriptsize(±2.77)} \\

    LLaMA 3.1-8B
      & 56.70 {\scriptsize(±3.07)}
      & 65.13 {\scriptsize(±2.94)}
      & 48.28 {\scriptsize(±3.09)}
      & 64.75 {\scriptsize(±2.96)}
      & 67.04 {\scriptsize(±2.91)}
      & 68.19 {\scriptsize(±2.88)}
      & \textbf{69.73} {\scriptsize(±2.84)} \\

    LLaMA 3.3-70B
      & 64.75 {\scriptsize(±2.96)}
      & 59.39 {\scriptsize(±3.04)}
      & 62.45 {\scriptsize(±3.00)}
      & 73.95 {\scriptsize(±2.72)}
      & 68.19 {\scriptsize(±2.88)}
      & 70.11 {\scriptsize(±2.80)}
      & \textbf{74.03} {\scriptsize(±2.78)} \\

    Claude 3.5 Haiku
      & 60.54 {\scriptsize(±3.02)}
      & 61.69 {\scriptsize(±3.01)}
      & 70.11 {\scriptsize(±2.83)}
      & 72.80 {\scriptsize(±2.76)}
      & 72.03 {\scriptsize(±2.78)}
      & 70.49 {\scriptsize(±2.83)}
      & \textbf{73.56} {\scriptsize(±2.73)} \\

    GPT-o3 Mini
      & 58.24 {\scriptsize(±3.05)}
      & 61.30 {\scriptsize(±3.02)}
      & \textbf{72.41} {\scriptsize(±2.77)}
      & 64.37 {\scriptsize(±2.97)}
      & 63.60 {\scriptsize(±2.92)}
      & 68.20 {\scriptsize(±2.88)}
      & 69.73 {\scriptsize(±2.84)} \\

    DeepSeek R1
      & 56.32 {\scriptsize(±3.07)}
      & 65.52 {\scriptsize(±2.94)}
      & 65.13 {\scriptsize(±2.94)}
      & 62.45 {\scriptsize(±3.00)}
      & 65.52 {\scriptsize(±2.91)}
      & 67.43 {\scriptsize(±2.90)}
      & \textbf{72.41} {\scriptsize(±2.77)} \\

    \midrule
    \rowcolor{gray!15}\multicolumn{8}{c}{\textbf{BioMedNLI*}} \\ \midrule
    Mistral 0.3-7B
      & 57.00 {\scriptsize(±3.53)}
      & 61.00 {\scriptsize(±3.45)}
      & 61.50 {\scriptsize(±3.44)}
      & 63.50 {\scriptsize(±3.42)}
      & 66.00 {\scriptsize(±3.35)}
      & 64.00 {\scriptsize(±3.39)}
      & \textbf{68.00} {\scriptsize(±3.29)} \\

    LLaMA 3.1-8B
      & 54.00 {\scriptsize(±3.53)}
      & 62.00 {\scriptsize(±3.44)}
      & 57.50 {\scriptsize(±3.51)}
      & 67.50 {\scriptsize(±3.37)}
      & 64.00 {\scriptsize(±3.39)}
      & 61.50 {\scriptsize(±3.44)}
      & \textbf{76.00} {\scriptsize(±3.04)} \\

    LLaMA 3.3-70B
      & 53.50 {\scriptsize(±3.53)}
      & 69.00 {\scriptsize(±3.30)}
      & 65.50 {\scriptsize(±3.39)}
      & 73.50 {\scriptsize(±3.16)}
      & 75.50 {\scriptsize(±3.06)}
      & 75.00 {\scriptsize(±3.06)}
      & \textbf{78.00} {\scriptsize(±2.94)} \\

    Claude 3.5 Haiku
      & 55.00 {\scriptsize(±3.53)}
      & 64.00 {\scriptsize(±3.43)}
      & 65.00 {\scriptsize(±3.39)}
      & 72.00 {\scriptsize(±3.13)}
      & 71.50 {\scriptsize(±3.19)}
      & 71.50 {\scriptsize(±3.19)}
      & \textbf{72.50} {\scriptsize(±3.11)} \\

    GPT-o3 Mini
      & 56.00 {\scriptsize(±3.53)}
      & 73.50 {\scriptsize(±3.16)}
      & 74.50 {\scriptsize(±3.11)}
      & 74.00 {\scriptsize(±3.13)}
      & 73.50 {\scriptsize(±3.12)}
      & 74.50 {\scriptsize(±3.08)}
      & \textbf{77.50} {\scriptsize(±2.95)} \\

    DeepSeek R1
      & 52.00 {\scriptsize(±3.53)}
      & \textbf{76.00} {\scriptsize(±3.04)}
      & 72.00 {\scriptsize(±3.22)}
      & 73.00 {\scriptsize(±3.18)}
      & 74.50 {\scriptsize(±3.08)}
      & 73.00 {\scriptsize(±3.14)}
      & 73.50 {\scriptsize(±3.16)} \\

    \midrule
    \rowcolor{gray!15}\multicolumn{8}{c}{\textbf{Dungeons and Dragons}} \\ \midrule
    Mistral 0.3-7B
      & 55.46 {\scriptsize(±4.56)}
      & 67.79 {\scriptsize(±3.83)}
      & 73.15 {\scriptsize(±3.63)}
      & 73.83 {\scriptsize(±3.60)}
      & 72.48 {\scriptsize(±3.66)}
      & 76.51 {\scriptsize(±3.47)}
      & \textbf{79.19} {\scriptsize(±3.33)} \\

    LLaMA 3.1-8B
      & 57.14 {\scriptsize(±4.54)}
      & 64.43 {\scriptsize(±3.92)}
      & 62.42 {\scriptsize(±3.97)}
      & 61.74 {\scriptsize(±3.98)}
      & 61.74 {\scriptsize(±3.98)}
      & 65.10 {\scriptsize(±3.90)}
      & \textbf{69.79} {\scriptsize(±3.76)} \\

    LLaMA 3.3-70B
      & 57.98 {\scriptsize(±4.52)}
      & 84.56 {\scriptsize(±2.96)}
      & 87.92 {\scriptsize(±2.67)}
      & 89.26 {\scriptsize(±2.54)}
      & 89.93 {\scriptsize(±2.47)}
      & 90.60 {\scriptsize(±2.39)}
      & \textbf{90.60} {\scriptsize(±2.39)} \\

    Claude 3.5 Haiku
      & 57.14 {\scriptsize(±4.54)}
      & 63.76 {\scriptsize(±3.94)}
      & 75.84 {\scriptsize(±3.51)}
      & \textbf{87.25} {\scriptsize(±2.73)}
      & 84.56 {\scriptsize(±2.96)}
      & 79.87 {\scriptsize(±3.29)}
      & 86.58 {\scriptsize(±2.79)} \\

    GPT-o3 Mini
      & 79.83 {\scriptsize(±3.68)}
      & 88.59 {\scriptsize(±2.60)}
      & 89.93 {\scriptsize(±2.47)}
      & 91.94 {\scriptsize(±2.06)}
      & 92.62 {\scriptsize(±2.14)}
      & 93.95 {\scriptsize(±2.04)}
      & \textbf{95.30} {\scriptsize(±1.73)} \\

    DeepSeek R1
      & 67.23 {\scriptsize(±4.30)}
      & 92.44 {\scriptsize(±2.42)}
      & 91.95 {\scriptsize(±2.23)}
      & 93.29 {\scriptsize(±2.05)}
      & 94.63 {\scriptsize(±2.14)}
      & 92.62 {\scriptsize(±2.14)}
      & \textbf{95.30} {\scriptsize(±1.74)} \\

    \bottomrule
  \end{tabular}
  \caption{Accuracy (with standard error) of \textsc{LogT} variants and baselines across four benchmarks. “ND” = No Document; “D” = Document; “LC” = Logic-based Context; “SGC” = Symbolic-Graph Context. Best per-row values are in \textbf{bold}.}
  \label{tab:main-results}
\end{table*}

\section{Results and Discussion}

\textbf{Finding 1: \textsc{LogT} consistently improves performance across all models, with smaller models exhibiting the most substantial gains.}  \textsc{LogT} significantly outperforms standard modes of prompting on all benchmarks, improving performance by an average of +4.41\% compared to the best baseline and +11.82\% compared to the average of all baselines (Table \ref{tab:main-results}). Promisingly, the largest gains are on the (initially) under-performing models. In ContractNLI, Mistral-7B and LLaMA-8B benefit the most, improving from 45.20\% and 44.30\% to 53.40\% (+8.2\%) and 50.40\% (+6.1\%), respectively. We see similar trends for these two models on other benchmarks. On average, Mistral-7B and LLaMA-8B outperform the (next) best and the average of all baselines by 6.73\% and 10.38\%, respectively.





\begin{figure*}[t]
  \centering
  \includegraphics[width=0.78\textwidth]{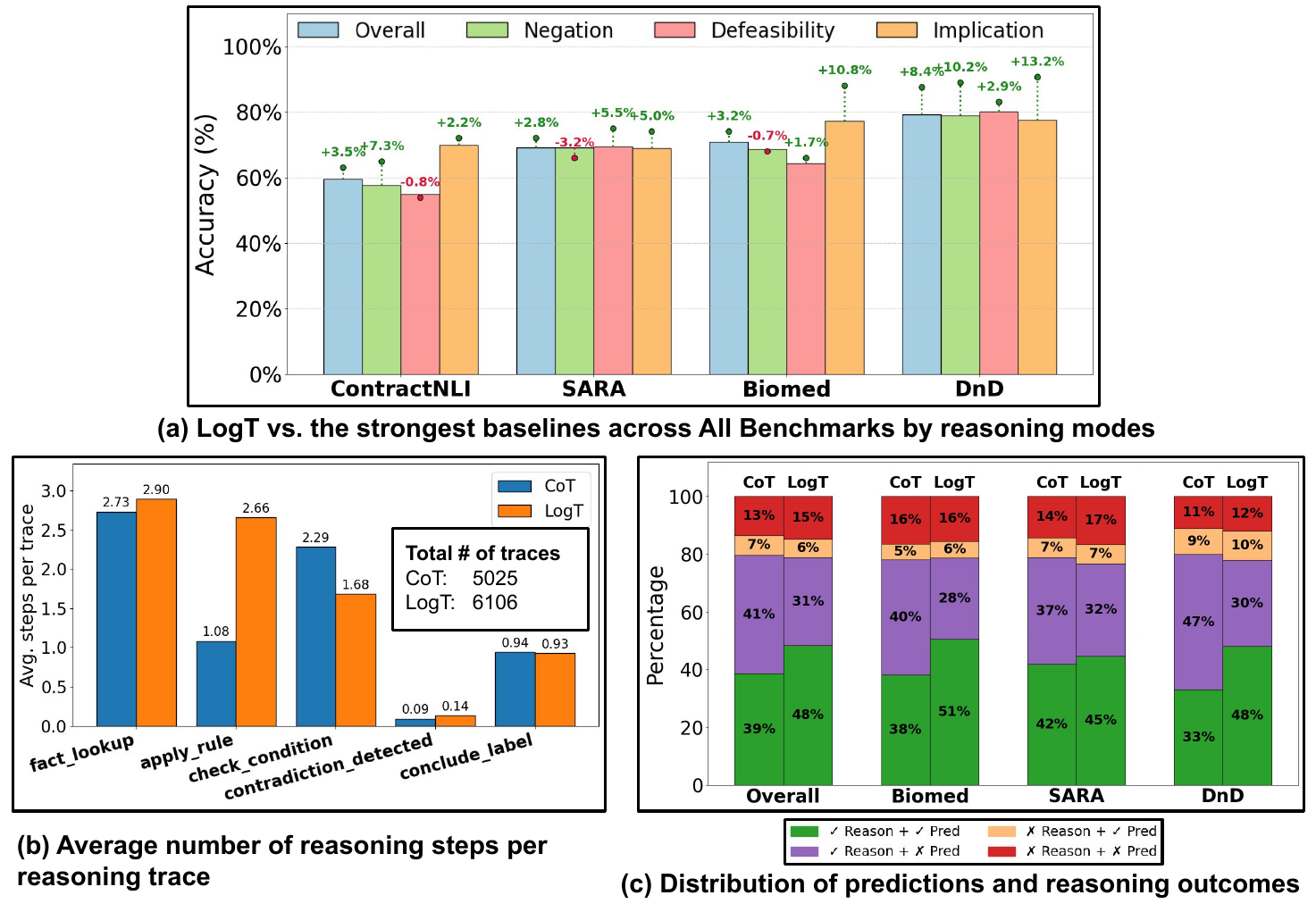}
  \caption{Performance evaluation of \textsc{LogT}: (a) shows an accuracy comparison between \textsc{LogT} (indicated by markers) and the strongest baselines (bars) across four benchmarks; (b) details the average number of reasoning steps per trace for both \textsc{LogT} and the CoT baseline; (c) displays the four outcome distributions for \textsc{LogT} and CoT, classified by whether the reasoning and final prediction were correct or incorrect. }
  \label{fig:results}
\end{figure*}

\textbf{Finding 2: \textsc{LogT} improves the ability of models across all three reasoning modes}. 
As presented in Figure \ref{fig:results} (a), we compared the performance of \textsc{LogT} to the \textit{best} performance among all prompting baselines (excluding \textsc{LogT w/o SGC} and \textsc{LogT w/o LC}). \textsc{LogT} consistently improves the performance of models across all three reasoning modes (\textit{negation, implication}, and \textit{defeasible}). The gains are most pronounced and consistent in the implication reasoning mode, where \textsc{LogT} boosts performance by +2.2\% on ContractNLI, +5.0\% on SARA, +10.8\% on Biomed, and a remarkable +13.2\% on DnD. Although, the impact of \textsc{LogT} generally improves the performance across all reasoning modes, we can see small performance drop on some areas, including performance drops of -0.8\% on ContractNLI (Defeasible), -3.2\% on SARA (Negation), and -0.7\% on BiomedNLI (Negation). This indicates that while \textsc{LogT} provides a strong overall benefit, its efficacy varies by the specific domains, presenting a clear direction for future refinement. We also provide the same analysis in Appendix H where we compare \textsc{LogT} performance against the average performance of all baselines instead of the best one. Extra analysis shows that \textsc{LogT} improves the performance in \textit{all} reasoning modes and benchmarks.



\textbf{Finding 3: Due to its explicit symbolic grounding, \textsc{LogT} elicits approximately 21.5\% more reasoning steps per example compared to standard CoT methods.} We compare the correctness of the reasoning trace between CoT and \textsc{LogT}. Figure \ref{fig:results} (b) presents statistics and a breakdown of the types of reasoning steps used by each method. On average, \textsc{LogT} generates 21.5\% more reasoning steps compared to Chain-of-Thought (CoT), despite using the same prompt instructions. This increase reflects a shift in the reasoning process rather than mere verbosity. Notably, the number of \textit{apply\_rule} steps increases substantially—from 1.08 in CoT to 2.66 in \textsc{LogT} demonstrating that \textsc{LogT} reasoning traces emphasize explicit rule-based deduction. Similarly, \textit{fact\_lookup} steps increase from 2.73 to 2.90, indicating that \textsc{LogT} encourages models to retrieve and verify factual information more actively than standard CoT. We provide qualitative evidence that LLMs utilized the provided neuro-symbolic contexts, especially on \textit{apply\_rule} and \textit{fact\_lookup} types, in Appendix H.


\textbf{Finding 4: \textsc{LogT} reasoning traces are more robust than CoT.} As shown in Figure \ref{fig:results} (c), the bar chart reveals that \textsc{LogT} improves the quality of reasoning traces. When the model produces correct reasoning, \textsc{LogT} leads to more accurate predictions (48\%), compared to only (39\%) for CoT. Moreover, \textsc{LogT} reduces the number of false negative cases, where correct reasoning leads to incorrect predictions compared to CoT. False negative cases reduce from 41\% (CoT) to 28\% (\textsc{LogT}). Hence, \textsc{LogT} not only produces more robust reasoning trace but also enhances its alignment with final predictions, resulting in notable accuracy gains (54\% vs. 46\% on average). The same trend as the overall can also be seen across all three benchmarks. We also provide confusion matrices to show the instance-level analysis in Appendix H.

\textbf{Finding 5 (ablation study): Both symbolic and logic-based contexts improve performance, but logic-based context contribute the most improvement.} We compare the improvement of \textit{LogT (SGC)} and \textit{LogT (LC)} with \textit{CoT} as presented in Table \ref{tab:main-results}. On average, LC improves accuracy by +2.4\% relative to CoT, whereas adding only the SGC yields a much smaller gain of +0.5\%. When both contexts are used together (\textsc{LogT (Full)}), the average improvement is boosted up to +7.4\%. The main exceptions appear in \textsc{BioMedNLI} and \textsc{Dungeons \& Dragons}, where six cases of SGC alone outperforms LC.


\section{Related Work}


\paragraph{High-Assurance Reasoning}

The integration of neural networks with symbolic reasoning is an important agenda in AI, with foundational neuro-symbolic systems developed in computer vision, rule learning, and knowledge representation \cite{andreas2016neural, hitzler2022neuro, hatzilygeroudis2004neuro, NEURIPS2019_c20a7ce2}. Contemporary methods within NLP typically take advantage of an LLM to translate natural language into an intermediate symbolic representation, such as program code or logical expressions \cite{nye2021improving, lyu2023faithful, pan2023logic}. However, existing methods typically focus on general logical formulations, neglecting specific high-assurance or non-monotonic reasoning scenarios such as defeasible logic. 
In contrast, our work builds on this neurosymbolic tradition but is primarily focused on high-assurance reasoning.

\paragraph{Natural Language Inference}

While Natural Language Inference (NLI) is a longstanding area of research \cite{cooper1994fracas}, it has taken on new importance in an era of transformer-based models. Even predating LLMs, recent work has taken on inference tasks of increasingly difficult and complexity \cite{bowman2015large,storks2019recent,welleck2018dialogue}, with a range of papers exploring inferences tasks that exhibit ambiguity, plausibility, and commonsense reasoning, for which ordinary entailment is insufficient \cite{nie2020can, meissner2021embracing}. Following this trend, NLI tasks both in domain-specific contexts such as law and health \cite{koreeda-manning-2021-contractnli-dataset, DBLP:conf/kdd/HolzenbergerBD20,romanov2018lessons, jullien-etal-2024-semeval,kazemi2023boardgameqa}, and those exploring non-monotonic reasoning, have been proposed but are still limited in scope, and do not involve formal evaluation of reasoning traces \cite{madaan2021could, rudinger2020thinking, yanaka2019can, DBLP:conf/aaai/ZhangJG25, DBLP:journals/corr/abs-2506-22385, gubelmann2024capturing}. 

\paragraph{Prompt Engineering in LLMs}

Prompt engineering and optimization has been an active area of research for getting the most of out LLMs, especially on complex problems requiring multi-step reasoning. In-context learning has demonstrated remarkable success, enabling models to tackle various tasks effectively with minimal examples, without updating model parameters \cite{brown2020language}. Chain-of-Thought (CoT), surveyed by \citet{yu2023towards,xia2024beyond,chen2025towards}, and its variants \cite{wei2022chain,wang2022self,long2023large, yao2023tree} are promising, but the lack of a symbolic infrastructure in CoT reasoning traces do not make them amenable to verification. To the best of our knowledge, the use of a dual context that involves knowledge and logic is novel. 



\bibliography{aaai2026}

\begin{thebibliography}{55}
\providecommand{\natexlab}[1]{#1}

\bibitem[{Andreas et~al.(2016)Andreas, Rohrbach, Darrell, and Klein}]{andreas2016neural}
Andreas, J.; Rohrbach, M.; Darrell, T.; and Klein, D. 2016.
\newblock Neural module networks.
\newblock In \emph{Proceedings of the IEEE conference on computer vision and pattern recognition}, 39--48.

\bibitem[{Blair-Stanek, Holzenberger, and Van~Durme(2023)}]{blair2023can}
Blair-Stanek, A.; Holzenberger, N.; and Van~Durme, B. 2023.
\newblock Can gpt-3 perform statutory reasoning?
\newblock In \emph{Proceedings of the Nineteenth International Conference on Artificial Intelligence and Law}, 22--31.

\bibitem[{Bowman et~al.(2015)Bowman, Angeli, Potts, and Manning}]{bowman2015large}
Bowman, S.~R.; Angeli, G.; Potts, C.; and Manning, C.~D. 2015.
\newblock A large annotated corpus for learning natural language inference.
\newblock \emph{arXiv preprint arXiv:1508.05326}.

\bibitem[{Brown et~al.(2020)Brown, Mann, Ryder, Subbiah, Kaplan, Dhariwal, Neelakantan, Shyam, Sastry, Askell et~al.}]{brown2020language}
Brown, T.; Mann, B.; Ryder, N.; Subbiah, M.; Kaplan, J.~D.; Dhariwal, P.; Neelakantan, A.; Shyam, P.; Sastry, G.; Askell, A.; et~al. 2020.
\newblock Language models are few-shot learners.
\newblock \emph{Advances in neural information processing systems}, 33: 1877--1901.

\bibitem[{Chen et~al.(2025)Chen, Qin, Liu, Peng, Guan, Wang, Hu, Zhou, Gao, and Che}]{chen2025towards}
Chen, Q.; Qin, L.; Liu, J.; Peng, D.; Guan, J.; Wang, P.; Hu, M.; Zhou, Y.; Gao, T.; and Che, W. 2025.
\newblock Towards reasoning era: A survey of long chain-of-thought for reasoning large language models.
\newblock \emph{arXiv preprint arXiv:2503.09567}.

\bibitem[{Cooper et~al.(1994)Cooper, Crouch, Van~Eijck, Fox, Van~Genabith, Jaspers, Kamp, Pinkal, Poesio, Pulman et~al.}]{cooper1994fracas}
Cooper, R.; Crouch, R.; Van~Eijck, J.; Fox, C.; Van~Genabith, J.; Jaspers, J.; Kamp, H.; Pinkal, M.; Poesio, M.; Pulman, S.; et~al. 1994.
\newblock FraCaS--a framework for computational semantics.
\newblock \emph{Deliverable D6}.

\bibitem[{Dhuliawala et~al.(2024)Dhuliawala, Komeili, Xu, Raileanu, Li, Celikyilmaz, and Weston}]{DBLP:conf/acl/DhuliawalaKXRLC24}
Dhuliawala, S.; Komeili, M.; Xu, J.; Raileanu, R.; Li, X.; Celikyilmaz, A.; and Weston, J. 2024.
\newblock Chain-of-Verification Reduces Hallucination in Large Language Models.
\newblock In Ku, L.; Martins, A.; and Srikumar, V., eds., \emph{Findings of the Association for Computational Linguistics, {ACL} 2024, Bangkok, Thailand and virtual meeting, August 11-16, 2024}, 3563--3578. Association for Computational Linguistics.

\bibitem[{Gao et~al.(2023)Gao, Madaan, Zhou, Alon, Liu, Yang, Callan, and Neubig}]{pmlr-v202-gao23f}
Gao, L.; Madaan, A.; Zhou, S.; Alon, U.; Liu, P.; Yang, Y.; Callan, J.; and Neubig, G. 2023.
\newblock {PAL}: Program-aided Language Models.
\newblock In Krause, A.; Brunskill, E.; Cho, K.; Engelhardt, B.; Sabato, S.; and Scarlett, J., eds., \emph{Proceedings of the 40th International Conference on Machine Learning}, volume 202 of \emph{Proceedings of Machine Learning Research}, 10764--10799. PMLR.

\bibitem[{Groen and Patel(2014)}]{groen2014relationship}
Groen, G.~J.; and Patel, V.~L. 2014.
\newblock The relationship between comprehension and reasoning in medical expertise.
\newblock In \emph{The nature of expertise}, 287--310. Psychology Press.

\bibitem[{Grosof et~al.(2023)Grosof, Kifer, Swift, Fodor, and Bloomfield}]{grosof2023ergo}
Grosof, B.; Kifer, M.; Swift, T.; Fodor, P.; and Bloomfield, J. 2023.
\newblock Ergo: a quest for declarativity in logic programming.
\newblock In \emph{Prolog: The Next 50 Years}, 224--236. Springer.

\bibitem[{Grosof, Kifer, and Fodor(2017)}]{grosof2017rulelog}
Grosof, B.~N.; Kifer, M.; and Fodor, P. 2017.
\newblock Rulelog: Highly Expressive Semantic Rules with Scalable Deep Reasoning.
\newblock In \emph{RuleML+ RR (Supplement)}.

\bibitem[{Gu et~al.(2024)Gu, Jiang, Shi, Tan, Zhai, Xu, Li, Shen, Ma, Liu et~al.}]{gu2024survey}
Gu, J.; Jiang, X.; Shi, Z.; Tan, H.; Zhai, X.; Xu, C.; Li, W.; Shen, Y.; Ma, S.; Liu, H.; et~al. 2024.
\newblock A survey on llm-as-a-judge.
\newblock \emph{arXiv preprint arXiv:2411.15594}.

\bibitem[{Gubelmann et~al.(2024)Gubelmann, Katis, Niklaus, and Handschuh}]{gubelmann2024capturing}
Gubelmann, R.; Katis, I.; Niklaus, C.; and Handschuh, S. 2024.
\newblock Capturing the varieties of natural language inference: A systematic survey of existing datasets and two novel benchmarks.
\newblock \emph{Journal of Logic, Language and Information}, 33(1): 21--48.

\bibitem[{Hatzilygeroudis and Prentzas(2004)}]{hatzilygeroudis2004neuro}
Hatzilygeroudis, I.; and Prentzas, J. 2004.
\newblock Neuro-symbolic approaches for knowledge representation in expert systems.
\newblock \emph{International Journal of Hybrid Intelligent Systems}, 1(3-4): 111--126.

\bibitem[{Hitzler et~al.(2022)Hitzler, Eberhart, Ebrahimi, Sarker, and Zhou}]{hitzler2022neuro}
Hitzler, P.; Eberhart, A.; Ebrahimi, M.; Sarker, M.~K.; and Zhou, L. 2022.
\newblock Neuro-symbolic approaches in artificial intelligence.
\newblock \emph{National Science Review}, 9(6): nwac035.

\bibitem[{Holzenberger, Blair{-}Stanek, and Durme(2020)}]{DBLP:conf/kdd/HolzenbergerBD20}
Holzenberger, N.; Blair{-}Stanek, A.; and Durme, B.~V. 2020.
\newblock A Dataset for Statutory Reasoning in Tax Law Entailment and Question Answering.
\newblock In Aletras, N.; Androutsopoulos, I.; Barrett, L.; Meyers, A.; and Preotiuc{-}Pietro, D., eds., \emph{Proceedings of the Natural Legal Language Processing Workshop 2020 co-located with the 26th {ACM} {SIGKDD} International Conference on Knowledge Discovery {\&} Data Mining {(KDD} 2020), Virtual Workshop, August 24, 2020}, volume 2645 of \emph{{CEUR} Workshop Proceedings}, 31--38. CEUR-WS.org.

\bibitem[{Holzenberger and Van~Durme(2023)}]{holzenberger2023connecting}
Holzenberger, N.; and Van~Durme, B. 2023.
\newblock Connecting symbolic statutory reasoning with legal information extraction.
\newblock In \emph{Proceedings of the Natural Legal Language Processing Workshop 2023}, 113--131. Association for Computational Linguistics.

\bibitem[{Hudson and Manning(2019)}]{NEURIPS2019_c20a7ce2}
Hudson, D.; and Manning, C.~D. 2019.
\newblock Learning by Abstraction: The Neural State Machine.
\newblock In Wallach, H.; Larochelle, H.; Beygelzimer, A.; d\textquotesingle Alch\'{e}-Buc, F.; Fox, E.; and Garnett, R., eds., \emph{Advances in Neural Information Processing Systems}, volume~32. Curran Associates, Inc.

\bibitem[{Jullien, Valentino, and Freitas(2024)}]{jullien-etal-2024-semeval}
Jullien, M.; Valentino, M.; and Freitas, A. 2024.
\newblock {S}em{E}val-2024 Task 2: Safe Biomedical Natural Language Inference for Clinical Trials.
\newblock In Ojha, A.~K.; Do{\u{g}}ru{\"o}z, A.~S.; Tayyar~Madabushi, H.; Da~San~Martino, G.; Rosenthal, S.; and Ros{\'a}, A., eds., \emph{Proceedings of the 18th International Workshop on Semantic Evaluation (SemEval-2024)}, 1947--1962. Mexico City, Mexico: Association for Computational Linguistics.

\bibitem[{Kazemi et~al.(2023)Kazemi, Yuan, Bhatia, Kim, Xu, Imbrasaite, and Ramachandran}]{kazemi2023boardgameqa}
Kazemi, M.; Yuan, Q.; Bhatia, D.; Kim, N.; Xu, X.; Imbrasaite, V.; and Ramachandran, D. 2023.
\newblock Boardgameqa: A dataset for natural language reasoning with contradictory information.
\newblock \emph{Advances in Neural Information Processing Systems}, 36: 39052--39074.

\bibitem[{Koreeda and Manning(2021)}]{koreeda-manning-2021-contractnli-dataset}
Koreeda, Y.; and Manning, C. 2021.
\newblock {C}ontract{NLI}: A Dataset for Document-level Natural Language Inference for Contracts.
\newblock In \emph{Findings of the Association for Computational Linguistics: EMNLP 2021}, 1907--1919. Punta Cana, Dominican Republic: Association for Computational Linguistics.

\bibitem[{Lanham et~al.(2023)Lanham, Chen, Radhakrishnan, Steiner, Denison, Hernandez, Li, Durmus, Hubinger, Kernion et~al.}]{lanham2023measuring}
Lanham, T.; Chen, A.; Radhakrishnan, A.; Steiner, B.; Denison, C.; Hernandez, D.; Li, D.; Durmus, E.; Hubinger, E.; Kernion, J.; et~al. 2023.
\newblock Measuring faithfulness in chain-of-thought reasoning.
\newblock \emph{arXiv preprint arXiv:2307.13702}.

\bibitem[{Li et~al.(2023{\natexlab{a}})Li, Shao, Xie, Sheng, Zheng, Gonzalez, Stoica, Ma, and Zhang}]{li2023long}
Li, D.; Shao, R.; Xie, A.; Sheng, Y.; Zheng, L.; Gonzalez, J.; Stoica, I.; Ma, X.; and Zhang, H. 2023{\natexlab{a}}.
\newblock How long can context length of open-source llms truly promise?
\newblock In \emph{NeurIPS 2023 Workshop on Instruction Tuning and Instruction Following}.

\bibitem[{Li et~al.(2023{\natexlab{b}})Li, Wang, Zheng, and Zhang}]{li2023loogle}
Li, J.; Wang, M.; Zheng, Z.; and Zhang, M. 2023{\natexlab{b}}.
\newblock Loogle: Can long-context language models understand long contexts?
\newblock \emph{arXiv preprint arXiv:2311.04939}.

\bibitem[{Li, Miao, and Li(2024)}]{li2024simple}
Li, M.; Miao, S.; and Li, P. 2024.
\newblock Simple is effective: The roles of graphs and large language models in knowledge-graph-based retrieval-augmented generation.
\newblock \emph{arXiv preprint arXiv:2410.20724}.

\bibitem[{Long(2023)}]{long2023large}
Long, J. 2023.
\newblock Large language model guided tree-of-thought.
\newblock \emph{arXiv preprint arXiv:2305.08291}.

\bibitem[{Lyu et~al.(2023)Lyu, Havaldar, Stein, Zhang, Rao, Wong, Apidianaki, and Callison-Burch}]{lyu2023faithful}
Lyu, Q.; Havaldar, S.; Stein, A.; Zhang, L.; Rao, D.; Wong, E.; Apidianaki, M.; and Callison-Burch, C. 2023.
\newblock Faithful chain-of-thought reasoning.
\newblock In \emph{The 13th International Joint Conference on Natural Language Processing and the 3rd Conference of the Asia-Pacific Chapter of the Association for Computational Linguistics (IJCNLP-AACL 2023)}.

\bibitem[{Madaan et~al.(2021)Madaan, Rajagopal, Tandon, Yang, and Hovy}]{madaan2021could}
Madaan, A.; Rajagopal, D.; Tandon, N.; Yang, Y.; and Hovy, E. 2021.
\newblock Could you give me a hint? generating inference graphs for defeasible reasoning.
\newblock \emph{arXiv preprint arXiv:2105.05418}.

\bibitem[{Meissner et~al.(2021)Meissner, Thumwanit, Sugawara, and Aizawa}]{meissner2021embracing}
Meissner, J.~M.; Thumwanit, N.; Sugawara, S.; and Aizawa, A. 2021.
\newblock Embracing ambiguity: Shifting the training target of NLI models.
\newblock \emph{arXiv preprint arXiv:2106.03020}.

\bibitem[{Narendra, Shetty, and Ratnaparkhi(2024)}]{narendra2024enhancing}
Narendra, S.; Shetty, K.; and Ratnaparkhi, A. 2024.
\newblock Enhancing contract negotiations with LLM-based legal document comparison.
\newblock In \emph{Proceedings of the Natural Legal Language Processing Workshop 2024}, 143--153.

\bibitem[{Nie, Zhou, and Bansal(2020)}]{nie2020can}
Nie, Y.; Zhou, X.; and Bansal, M. 2020.
\newblock What can we learn from collective human opinions on natural language inference data?
\newblock \emph{arXiv preprint arXiv:2010.03532}.

\bibitem[{Nye et~al.(2021)Nye, Tessler, Tenenbaum, and Lake}]{nye2021improving}
Nye, M.; Tessler, M.; Tenenbaum, J.; and Lake, B.~M. 2021.
\newblock Improving coherence and consistency in neural sequence models with dual-system, neuro-symbolic reasoning.
\newblock \emph{Advances in Neural Information Processing Systems}, 34: 25192--25204.

\bibitem[{Pan et~al.(2023)Pan, Albalak, Wang, and Wang}]{pan2023logic}
Pan, L.; Albalak, A.; Wang, X.; and Wang, W.~Y. 2023.
\newblock Logic-lm: Empowering large language models with symbolic solvers for faithful logical reasoning.
\newblock \emph{arXiv preprint arXiv:2305.12295}.

\bibitem[{Peng et~al.(2024)Peng, Zhu, Liu, Bo, Shi, Hong, Zhang, and Tang}]{peng2024graph}
Peng, B.; Zhu, Y.; Liu, Y.; Bo, X.; Shi, H.; Hong, C.; Zhang, Y.; and Tang, S. 2024.
\newblock Graph retrieval-augmented generation: A survey.
\newblock \emph{arXiv preprint arXiv:2408.08921}.

\bibitem[{Romanov and Shivade()}]{romanov2018lessons}
Romanov, A.; and Shivade, C. ????
\newblock Lessons from Natural Language Inference in the Clinical Domain.

\bibitem[{Rudinger et~al.(2020)Rudinger, Shwartz, Hwang, Bhagavatula, Forbes, Le~Bras, Smith, and Choi}]{rudinger2020thinking}
Rudinger, R.; Shwartz, V.; Hwang, J.~D.; Bhagavatula, C.; Forbes, M.; Le~Bras, R.; Smith, N.~A.; and Choi, Y. 2020.
\newblock Thinking like a skeptic: Defeasible inference in natural language.
\newblock In \emph{Findings of the Association for Computational Linguistics: EMNLP 2020}, 4661--4675.

\bibitem[{Storks, Gao, and Chai(2019)}]{storks2019recent}
Storks, S.; Gao, Q.; and Chai, J.~Y. 2019.
\newblock Recent advances in natural language inference: A survey of benchmarks, resources, and approaches.
\newblock \emph{arXiv preprint arXiv:1904.01172}.

\bibitem[{Susskind(1986)}]{susskind1986expert}
Susskind, R.~E. 1986.
\newblock Expert systems in law: A jurisprudential approach to artificial intelligence and legal reasoning.
\newblock \emph{The modern law review}, 49(2): 168--194.

\bibitem[{Turpin et~al.(2023)Turpin, Michael, Perez, and Bowman}]{turpin2023language}
Turpin, M.; Michael, J.; Perez, E.; and Bowman, S. 2023.
\newblock Language models don't always say what they think: Unfaithful explanations in chain-of-thought prompting.
\newblock \emph{Advances in Neural Information Processing Systems}, 36: 74952--74965.

\bibitem[{Wang et~al.(2022)Wang, Wei, Schuurmans, Le, Chi, Narang, Chowdhery, and Zhou}]{wang2022self}
Wang, X.; Wei, J.; Schuurmans, D.; Le, Q.; Chi, E.; Narang, S.; Chowdhery, A.; and Zhou, D. 2022.
\newblock Self-consistency improves chain of thought reasoning in language models.
\newblock \emph{arXiv preprint arXiv:2203.11171}.

\bibitem[{Wei et~al.(2022{\natexlab{a}})Wei, Wang, Schuurmans, Bosma, Ichter, Xia, Chi, Le, and Zhou}]{DBLP:conf/nips/Wei0SBIXCLZ22}
Wei, J.; Wang, X.; Schuurmans, D.; Bosma, M.; Ichter, B.; Xia, F.; Chi, E.~H.; Le, Q.~V.; and Zhou, D. 2022{\natexlab{a}}.
\newblock Chain-of-Thought Prompting Elicits Reasoning in Large Language Models.
\newblock In Koyejo, S.; Mohamed, S.; Agarwal, A.; Belgrave, D.; Cho, K.; and Oh, A., eds., \emph{Advances in Neural Information Processing Systems 35: Annual Conference on Neural Information Processing Systems 2022, NeurIPS 2022, New Orleans, LA, USA, November 28 - December 9, 2022}.

\bibitem[{Wei et~al.(2022{\natexlab{b}})Wei, Wang, Schuurmans, Bosma, Xia, Chi, Le, Zhou et~al.}]{wei2022chain}
Wei, J.; Wang, X.; Schuurmans, D.; Bosma, M.; Xia, F.; Chi, E.; Le, Q.~V.; Zhou, D.; et~al. 2022{\natexlab{b}}.
\newblock Chain-of-thought prompting elicits reasoning in large language models.
\newblock \emph{Advances in neural information processing systems}, 35: 24824--24837.

\bibitem[{Welleck et~al.(2018)Welleck, Weston, Szlam, and Cho}]{welleck2018dialogue}
Welleck, S.; Weston, J.; Szlam, A.; and Cho, K. 2018.
\newblock Dialogue natural language inference.
\newblock \emph{arXiv preprint arXiv:1811.00671}.

\bibitem[{Xia et~al.(2024)Xia, Wang, Liu, Li, Yu, Chen, McAuley, and Li}]{xia2024beyond}
Xia, Y.; Wang, R.; Liu, X.; Li, M.; Yu, T.; Chen, X.; McAuley, J.; and Li, S. 2024.
\newblock Beyond chain-of-thought: A survey of chain-of-x paradigms for llms.
\newblock \emph{arXiv preprint arXiv:2404.15676}.

\bibitem[{Xiu, Xiao, and Liu(2022)}]{xiu2022logicnmr}
Xiu, Y.; Xiao, Z.; and Liu, Y. 2022.
\newblock LogicNMR: Probing the non-monotonic reasoning ability of pre-trained language models.
\newblock In \emph{Findings of the Association for Computational Linguistics: EMNLP 2022}, 3616--3626.

\bibitem[{Yanaka et~al.(2019)Yanaka, Mineshima, Bekki, Inui, Sekine, Abzianidze, and Bos}]{yanaka2019can}
Yanaka, H.; Mineshima, K.; Bekki, D.; Inui, K.; Sekine, S.; Abzianidze, L.; and Bos, J. 2019.
\newblock Can neural networks understand monotonicity reasoning?
\newblock \emph{arXiv preprint arXiv:1906.06448}.

\bibitem[{Yao et~al.(2024)Yao, Chen, Zou, Lu, Li, Zhang, Sang, Liu, Hendler, and Wang}]{yao-etal-2024-samples}
Yao, B.; Chen, G.; Zou, R.; Lu, Y.; Li, J.; Zhang, S.; Sang, Y.; Liu, S.; Hendler, J.; and Wang, D. 2024.
\newblock More Samples or More Prompts? Exploring Effective Few-Shot In-Context Learning for {LLM}s with In-Context Sampling.
\newblock In Duh, K.; Gomez, H.; and Bethard, S., eds., \emph{Findings of the Association for Computational Linguistics: NAACL 2024}, 1772--1790. Mexico City, Mexico: Association for Computational Linguistics.

\bibitem[{Yao et~al.(2023)Yao, Yu, Zhao, Shafran, Griffiths, Cao, and Narasimhan}]{yao2023tree}
Yao, S.; Yu, D.; Zhao, J.; Shafran, I.; Griffiths, T.; Cao, Y.; and Narasimhan, K. 2023.
\newblock Tree of thoughts: Deliberate problem solving with large language models.
\newblock \emph{Advances in neural information processing systems}, 36: 11809--11822.

\bibitem[{Yu et~al.(2023)Yu, He, Wu, Dai, and Chen}]{yu2023towards}
Yu, Z.; He, L.; Wu, Z.; Dai, X.; and Chen, J. 2023.
\newblock Towards better chain-of-thought prompting strategies: A survey.
\newblock \emph{arXiv preprint arXiv:2310.04959}.

\bibitem[{Yuan et~al.(2023)Yuan, Chen, Cui, Gao, Zou, Cheng, Ji, Liu, and Sun}]{yuan2023revisiting}
Yuan, L.; Chen, Y.; Cui, G.; Gao, H.; Zou, F.; Cheng, X.; Ji, H.; Liu, Z.; and Sun, M. 2023.
\newblock Revisiting out-of-distribution robustness in nlp: Benchmarks, analysis, and llms evaluations.
\newblock \emph{Advances in Neural Information Processing Systems}, 36: 58478--58507.

\bibitem[{Zhang et~al.(2024)Zhang, Du, Pang, Liu, Gao, and Lin}]{zhang2024chain}
Zhang, X.; Du, C.; Pang, T.; Liu, Q.; Gao, W.; and Lin, M. 2024.
\newblock Chain of preference optimization: Improving chain-of-thought reasoning in llms.
\newblock \emph{Advances in Neural Information Processing Systems}, 37: 333--356.

\bibitem[{Zhang, Jing, and Gogate(2025)}]{DBLP:conf/aaai/ZhangJG25}
Zhang, Y.; Jing, L.; and Gogate, V. 2025.
\newblock Defeasible Visual Entailment: Benchmark, Evaluator, and Reward-Driven Optimization.
\newblock In Walsh, T.; Shah, J.; and Kolter, Z., eds., \emph{AAAI-25, Sponsored by the Association for the Advancement of Artificial Intelligence, February 25 - March 4, 2025, Philadelphia, PA, {USA}}, 25976--25984. {AAAI} Press.

\bibitem[{Zhang et~al.(2025)Zhang, Sun, Guo, and Gogate}]{DBLP:journals/corr/abs-2506-22385}
Zhang, Y.; Sun, J.; Guo, Y.; and Gogate, V. 2025.
\newblock Can Video Large Multimodal Models Think Like Doubters-or Double-Down: {A} Study on Defeasible Video Entailment.
\newblock \emph{CoRR}, abs/2506.22385.

\bibitem[{Zheng et~al.(2023)Zheng, Chiang, Sheng, Zhuang, Wu, Zhuang, Lin, Li, Li, Xing et~al.}]{zheng2023judging}
Zheng, L.; Chiang, W.-L.; Sheng, Y.; Zhuang, S.; Wu, Z.; Zhuang, Y.; Lin, Z.; Li, Z.; Li, D.; Xing, E.; et~al. 2023.
\newblock Judging llm-as-a-judge with mt-bench and chatbot arena.
\newblock \emph{Advances in neural information processing systems}, 36: 46595--46623.

\bibitem[{Zhou et~al.(2021)Zhou, Khanna, Lee, Lin, Ho, Pujara, and Ren}]{zhou-etal-2021-rica}
Zhou, P.; Khanna, R.; Lee, S.; Lin, B.~Y.; Ho, D.; Pujara, J.; and Ren, X. 2021.
\newblock {RICA}: Evaluating Robust Inference Capabilities Based on Commonsense Axioms.
\newblock In Moens, M.-F.; Huang, X.; Specia, L.; and Yih, S. W.-t., eds., \emph{Proceedings of the 2021 Conference on Empirical Methods in Natural Language Processing}, 7560--7579. Online and Punta Cana, Dominican Republic: Association for Computational Linguistics.

\end{thebibliography}

\clearpage

\section*{Technical Appendix Contents}

This Technical Appendix, a supplement to the paper \textit{LOGicalThought: Logic-Based Ontological Grounding of LLMs for High-Assurance Reasoning}. This appendix consists of:

\begin{tabularx}{\linewidth}{@{}l X@{}}
\textbf{Appendix A:} & Model Summary \\
\textbf{Appendix B:} & Experiment Configuration \\
\textbf{Appendix C:} & Prompt Templates and Output Examples \\
\textbf{Appendix C.1:} & Benchmark Enhancement \\
\textbf{Appendix C.2:} & Symbolic Graph Context \\
\textbf{Appendix C.3:} & ErgoAI Program Generation \\
\textbf{Appendix C.4:} & Reasoning Trace Analysis \\
\textbf{Appendix C.5:} & Grounded LLM Evaluation \\
\textbf{Appendix D:} & Hypothesis Examples of Different Reasoning Modes \\
\textbf{Appendix E:} & ErgoAI Program Synthesis \\
\textbf{Appendix F:} & Benchmark Statistics \\
\textbf{Appendix G:} & Pilot Study Results and NLI \\
\textbf{Appendix H:} & Additional Evaluation Results \\
\textbf{Appendix H.1:} & LOGT Performance against the Average Performance of all Baselines \\
\textbf{Appendix H.2:} & Reasoning Trace Instance-level Analysis \\
\textbf{Appendix H.3:} & Reasoning Trace Qualitative Analysis \\
\end{tabularx}

\section{Appendix A - Model Summary}
\label{app:model-summary}

\begin{table*}[t!]
\centering
\small
\renewcommand{\arraystretch}{1.6}  
\begin{tabular}{@{}lllccl@{}}
\toprule
\textbf{Weight Type} & \textbf{Model} & \textbf{Platform (Provider)} & \textbf{Released} & \textbf{Training Cut-off} & \textbf{Parameters} \\
\midrule
\multirow{4}{*}{Open source}
& Mistral Instruct v0.3 & Together (Mistral) & May 2024 & Unknown & 7B \\
& LLaMA 3.1 & Together (Meta) & Jul 2024 & Dec 2023 & 8B \\
& LLaMA 3.3 & Together (Meta) & Dec 2024 & Dec 2023 & 70B \\
& DeepSeek R1 & Deepseek & Jan 2025 & Jul 2024 & 67B \\
\midrule
\multirow{2}{*}{Closed source}
& GPT-o3 Mini & OpenAI & Jul 2024 & Oct 2023 & ? \\
& Claude 3.5 Haiku & Anthropic & Oct 2024 & Jul 2024 & ? \\
\bottomrule
\end{tabular}
\caption{Overview of the models evaluated and utilized in case studies.}
\label{tab:model_overview}
\end{table*}

We present a summary of the models used in our experiments in Table \ref{tab:model_overview}.
Our models were selected in order to represent popular LLMs in use today while also covering a range of sizes and providers.
We use 4 open source models and 2 closed source models from the two promiment providers (OpenAI and Anthropic).
All models were accessed via OpenRouter's Python API\footnote{https://openrouter.ai/}.

\section{Appendix B - Experiment Configuration}
\label{app:exp-config}

To ensure the reproducibility of our experiments, this section provides the complete details of the language model configurations and the prompt templates used for our primary tasks.

\paragraph{Reproducibility Setup}
All symbolic reasoning querying was conducted locally on a MacBook Pro (Apple M1 Pro, CPU), with ErgoAI installed natively. This is specifically for compilability check and logic program execution. Additionally, all language model prompts and evaluation scripts are designed to be hardware-agnostic and can be executed on both the local machine (CPU) and Google Colab (CPU). This setup ensures accessibility and ease of replication without requiring GPU acceleration.

\subsection*{Language Models Configuration}
To ensure fair comparison and reproducibility, a consistent set of hyperparameters was used for all models listed in Appendix A across all experiments. This standardized approach was applied to all baseline generation strategies and \textsc{LogT} variants.

\paragraph{Prediction Generation.}
For all final prediction steps, we employed a deterministic configuration with temperature set to 0.0, a fixed seed of 42, and a maximum token limit of 4096. This setup was used for the \textit{Basic-ND}, \textit{Basic-D}, and \textit{Few-shot} baselines, as well as for the final prediction stage of both \textit{CoT} and all \textsc{LogT} variants.

\paragraph{Reasoning Generation.}
To encourage more elaborative intermediate reasoning, we increased the temperature slightly to 0.2 during the analysis-generation phase within the \textit{CoT} and \textsc{LogT} pipelines. However, the final prediction after this reasoning step still used temperature 0.0 for consistency.

\subsection*{Benchmark Enhancement and Hypothesis Generation}
To ensure a robust and balanced evaluation, the test sets were curated by generating new hypotheses for some benchmarks. This process was standardized to create consistent and reproducible datasets.

\paragraph{Hypothesis Enhancement Configuration.}
All supplementary hypotheses were generated using GPT-4o with a deterministic setup: temperature was set to 0.0, and the seed was fixed at 42. This configuration guarantees the generation of a static and perfectly reproducible dataset. Our main objective in this process was to ensure an even distribution of instances across the three core reasoning modes: Negation, Implication, and Defeasible.

\section{Appendix C - Prompt Templates}

This appendix provides a comprehensive overview of the prompt templates used throughout our study. These templates form the foundation of our experimental design, supporting various stages of the workflow including benchmark enhancement, context enrichment, reasoning trace generation, and grounded model evaluation. The prompts are designed to enable systematic probing of model reasoning capabilities, generalization, and alignment with symbolic or logical information. To streamline presentation, we consolidate all templates in this section and present them as visual examples in the following pages.

\textbf{Appendix C.1 Benchmark Enhancement for Non-monotonic Reasoning.} This category includes prompt templates used to generate controlled variations of benchmark hypotheses across different reasoning types. Specifically, we define 11 distinct formats that introduce transformations aligned with negation, causality, or defeasibility. The goal is to ensure a balanced and diverse evaluation set that challenges the model’s ability to reason under different conditions. For example, “Negation\_Flip” inverts the polarity of the hypothesis, while “Entailment\_Defeasibility” introduces an entailment that holds under normal but defeasible assumptions. Table~\ref{tab:c1-prompt-types} summarizes the full set of formats.

\begin{table}[h]
\centering
\small
\setlength{\tabcolsep}{8pt}
\renewcommand{\arraystretch}{1.6}
\begin{tabular}{@{}ll@{}}
\toprule
\textbf{Template Format Name} & \textbf{Reasoning Mode} \\
\midrule
Negation\_Flip & Negation \\
Negation\_Maintain & Negation \\
Causality\_Entailment & Implication \\
Causality\_Contradiction & Implication \\
NotMentioned\_Defeasibility & Defeasible \\
NotMentioned\_Negation & Negation \\
NotMentioned\_Causality & Implication \\
Entailment\_Defeasibility & Defeasible \\
Contradiction\_Defeasibility & Defeasible \\
Defeasibility\_Entailment & Defeasible \\
Defeasibility\_Contradiction & Defeasible \\
\bottomrule
\end{tabular}
\caption{List of 11 prompt templates categorized by reasoning mode: Negation, Implication, and Defeasible.}
\label{tab:c1-prompt-types}
\end{table}

\textbf{Appendix C.2 Symbolic Graph Context.} These prompts instruct the language model to extract symbolic triples from natural language hypotheses and contexts. Each triple captures a relation in the form of \textit{subject–predicate–object}, allowing us to construct a structured semantic graph. This symbolic representation is later used as additional context to enhance model reasoning. For instance, given a biomedical statement, the prompt may elicit triples like “(fever, indicates, infection),” providing an interpretable scaffold for downstream inference.

\textbf{Appendix C.3 ErgoAI Program Generation.} To support formal reasoning, we design prompts that generate lightweight logic programs using the ErgoAI language. These logic-based contexts capture conditional rules, exceptions, and defeasible logic extracted from the input. The generated programs serve as abstract representations of background knowledge, enabling structured inference. 

\textbf{Appendix C.4 Reasoning Traces.} This class of prompts elicits step-by-step explanations from language models. Rather than asking for direct answers, we guide the model to reason explicitly—identifying relevant facts, linking them, and drawing conclusions in a transparent manner. The output typically begins with a statement like “First, we note that...,” and proceeds through a logical narrative before arriving at a final prediction. These traces improve interpretability and allow for trace-level evaluation of model reasoning quality.

\textbf{Appendix C.5 Grounded LLM Calls.} Finally, we develop prompt templates that combine the original input with structured context—either symbolic graphs, logical-based contexts, or both—before querying the model for a prediction. These prompts allow us to test whether and how external structure improves performance. By controlling the form and source of the grounding, we can compare zero-shot, symbolic-only, logic-only, and combined configurations. This setup enables a systematic evaluation of grounding effectiveness under different model architectures and settings.

Together, these prompt templates define a modular and extensible interface between symbolic knowledge, logical reasoning, and language models. The full prompt examples for each category are provided as figures in the remainder of this appendix.

\clearpage
\subsection{Appendix C.1 Benchmark Enhancement for Non-monotonic Reasoning.}

\begin{figure}[h]
    \centering
    \includegraphics[width=\textwidth]{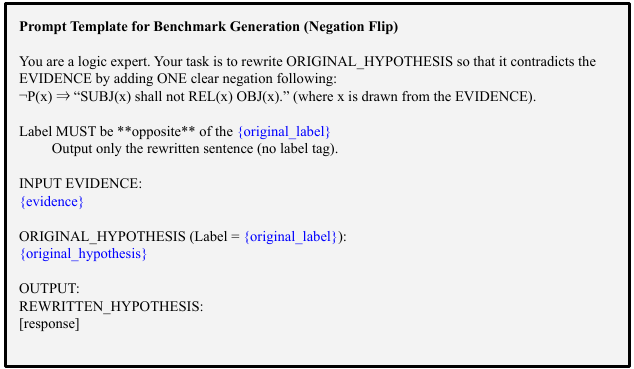}
    \label{fig:c1-negation-flip}
\end{figure}

\begin{figure*}[t!]
    \centering
    \includegraphics[width=0.95\textwidth]{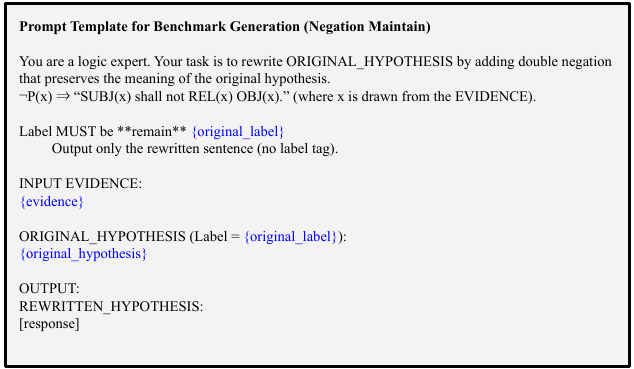}
    \label{fig:c1-negation-maintain}
\end{figure*}

\begin{figure*}[t!]
    \centering
    \includegraphics[width=0.95\textwidth]{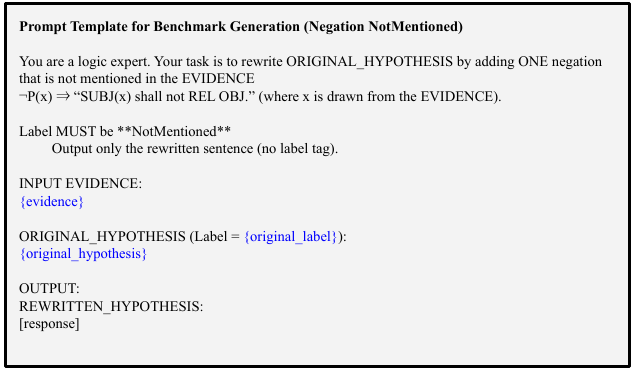}
    \label{fig:c1-negation-notmentioned}
\end{figure*}

\begin{figure*}[t!]
    \centering
    \includegraphics[width=0.95\textwidth]{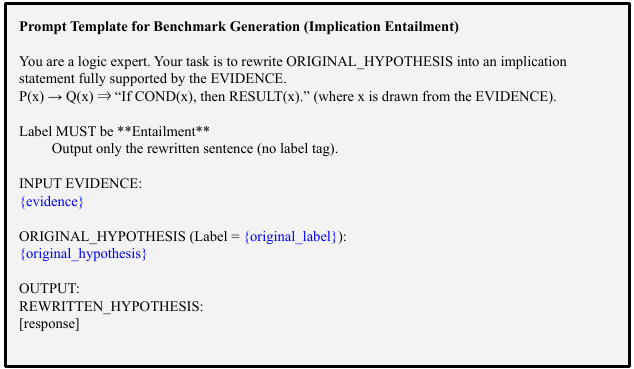}
    \label{fig:c1-implication-entailment}
\end{figure*}

\begin{figure*}[t!]
    \centering
    \includegraphics[width=0.95\textwidth]{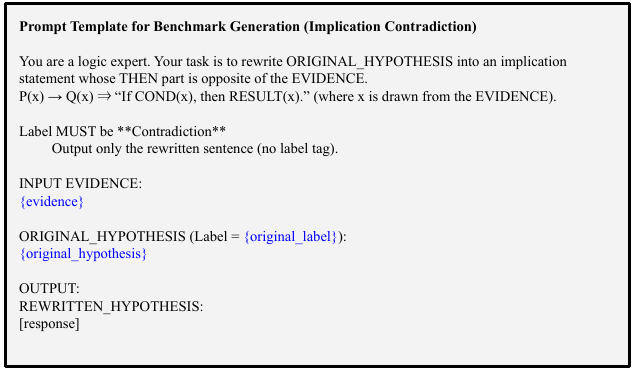}
    \label{fig:c1-implication-contradiction}
\end{figure*}

\begin{figure*}[t!]
    \centering
    \includegraphics[width=0.95\textwidth]{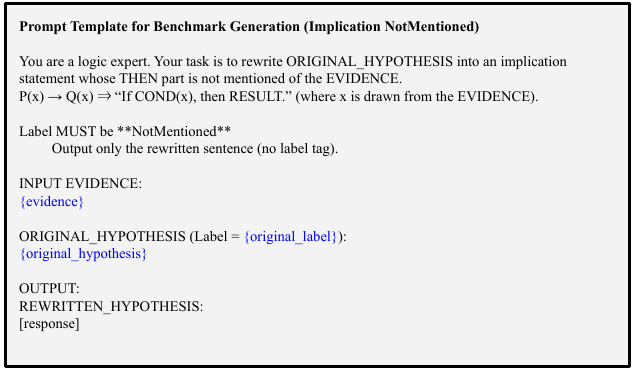}
    \label{fig:c1-implication-notmentioned}
\end{figure*}

\begin{figure*}[t!]
    \centering
    \includegraphics[width=0.95\textwidth]{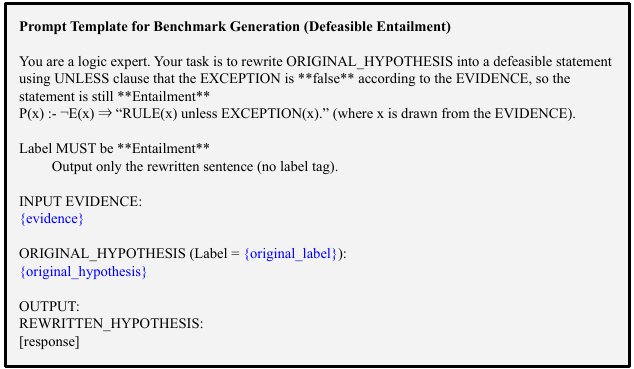}
    \label{fig:c1-defeasible-entailment}
\end{figure*}

\begin{figure*}[t!]
    \centering
    \includegraphics[width=0.95\textwidth]{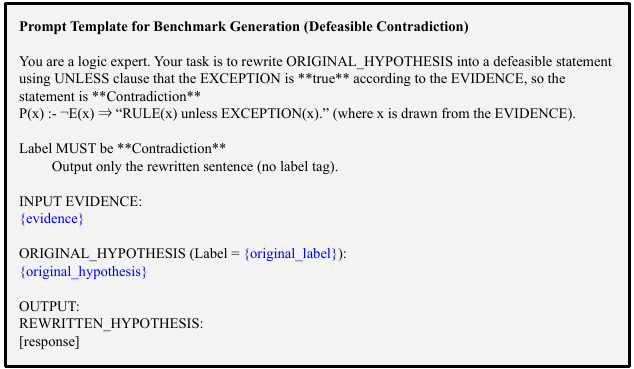}
    \label{fig:c1-defeasible-contradiction}
\end{figure*}

\begin{figure*}[t!]
    \centering
    \includegraphics[width=0.95\textwidth]{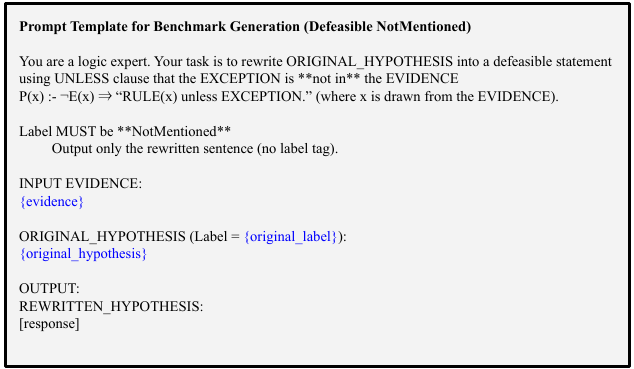}
    \label{fig:c1-defeasible-notmentioned-1}
\end{figure*}

\begin{figure*}[t!]
    \centering
    \includegraphics[width=0.95\textwidth]{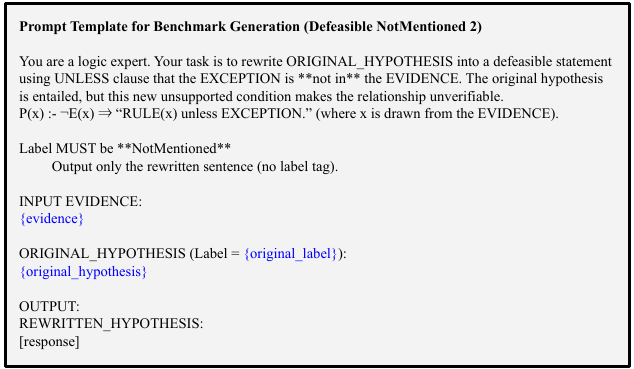}
    \label{fig:c1-defeasible-notmentioned-2}
\end{figure*}

\begin{figure*}[t!]
    \centering
    \includegraphics[width=0.95\textwidth]{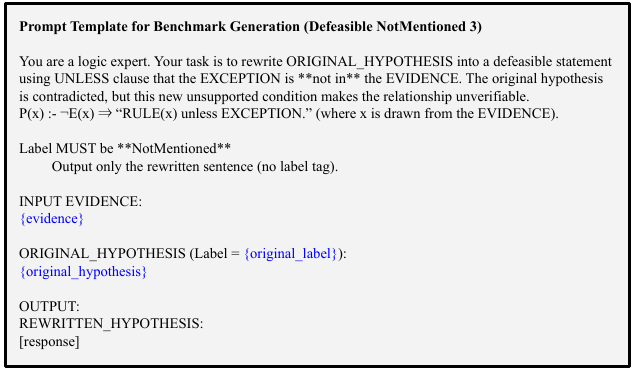}
    \label{fig:c1-defeasible-notmentioned-3}
\end{figure*}

\clearpage
\subsection{Appendix C.2 Prompt Template for Symbolic Graph Context.}

\begin{figure}[h]
    \centering
    \includegraphics[width=\textwidth]{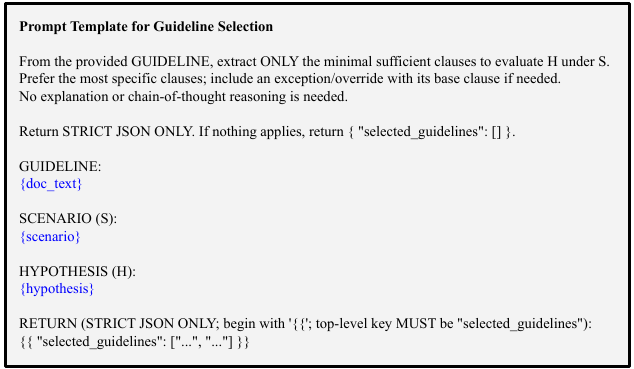}
    \label{fig:c2-symbolic-select}
\end{figure}

\begin{figure*}[!t]
    \centering
    \includegraphics[width=0.85\textwidth]{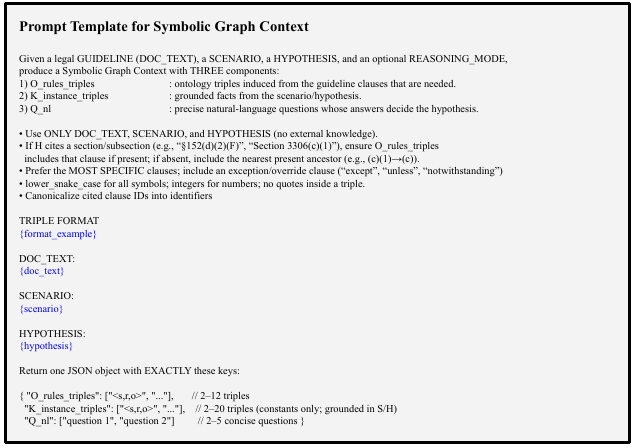}
    \label{fig:c2-ontology-generation}
\end{figure*}

\begin{figure*}[!t]
    \centering
    \includegraphics[width=0.85\textwidth]{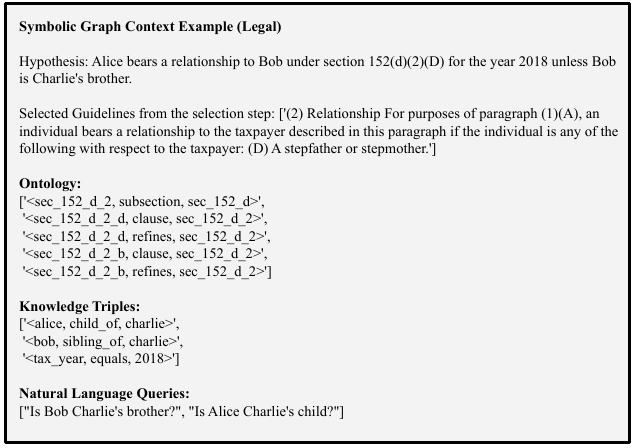}
    \label{fig:c2-symbolic-context-legal}
\end{figure*}

\begin{figure*}[!t]
    \centering
    \includegraphics[width=0.85\textwidth]{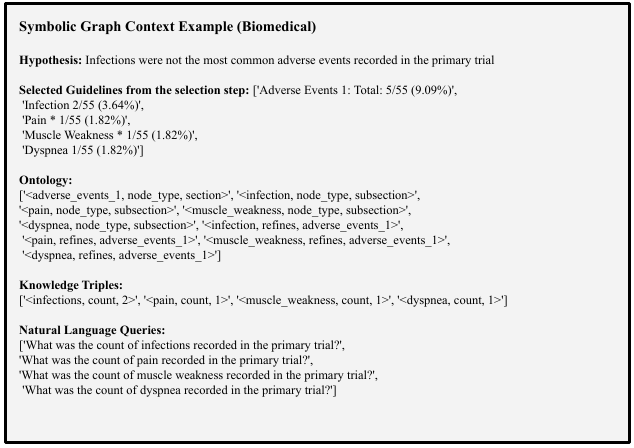}
    \label{fig:c2-symbolic-context-biomed}
\end{figure*}

\begin{figure*}[!t]
    \centering
    \includegraphics[width=0.85\textwidth]{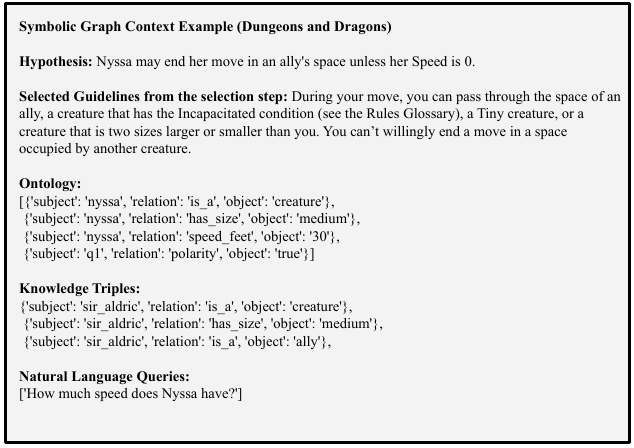}
    \label{fig:c2-symbolic-context-dnd}
\end{figure*}

\clearpage

\subsection{Appendix C.3 - Prompt Template for ErgoAI Program Generation.}

\begin{figure}[h]
    \centering
    \includegraphics[width=\textwidth]{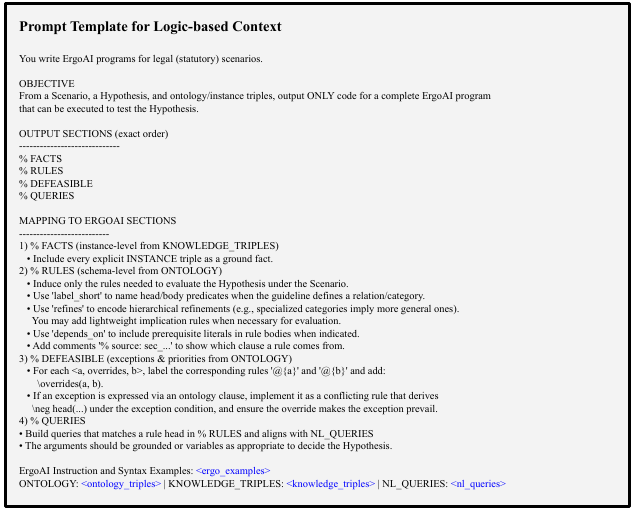}
    \label{fig:c3-ergo-gen}
\end{figure}

\begin{figure*}[!t]
    \centering
    \includegraphics[width=\textwidth]{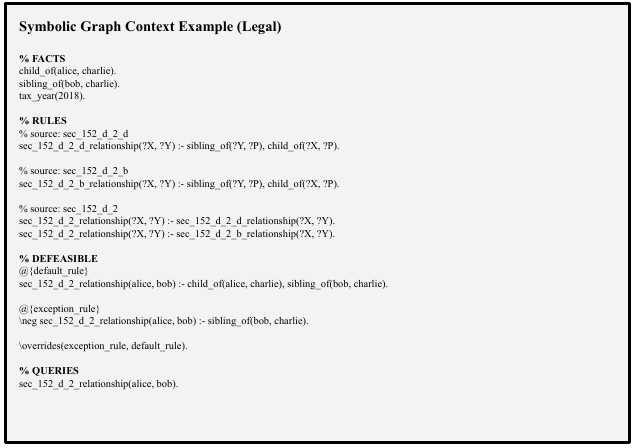}
    \label{fig:c3-ergo-program-legal}
\end{figure*}

\begin{figure*}[!t]
    \centering
    \includegraphics[width=\textwidth]{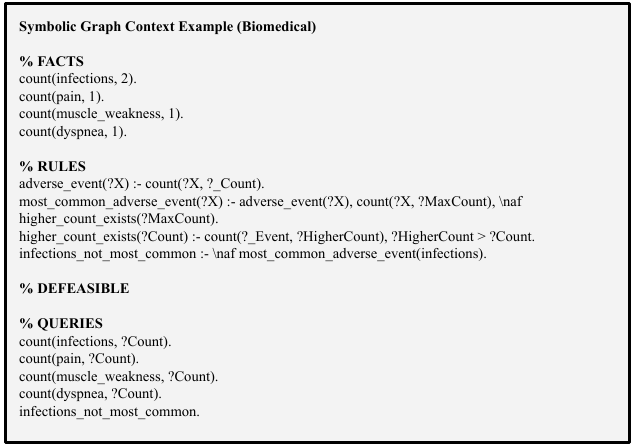}
    \label{fig:c3-ergo-program-biomed}
\end{figure*}

\begin{figure*}[!t]
    \centering
    \includegraphics[width=\textwidth]{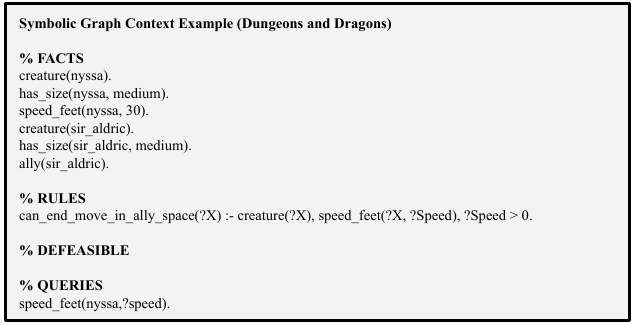}
    \label{fig:c3-ergo-program-dnd}
\end{figure*}

\clearpage
\subsection{Appendix C.4 - Prompt Template for Reasoning Trace.}

\begin{figure}[h]
    \centering
    \includegraphics[width=\textwidth]{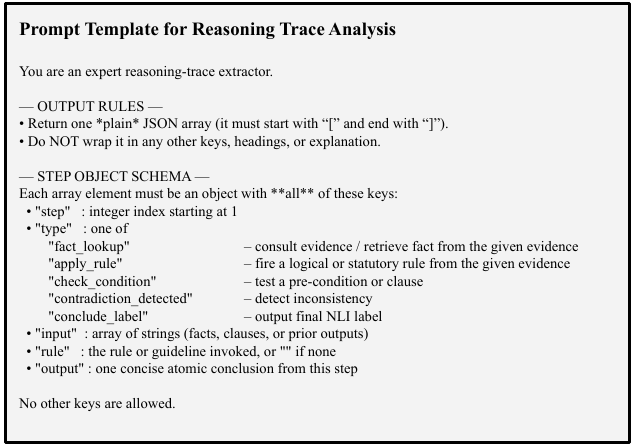}
    \label{fig:c3-ergo-gen}
\end{figure}

\begin{figure*}[!t]
    \centering
    \includegraphics[width=\textwidth]{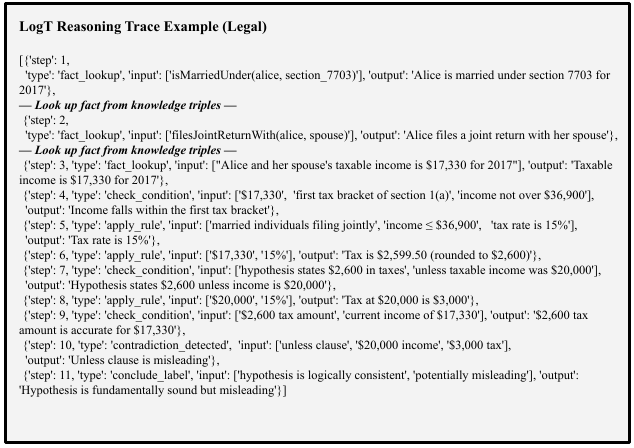}
\end{figure*}

\begin{figure*}[!t]
    \centering
    \includegraphics[width=\textwidth]{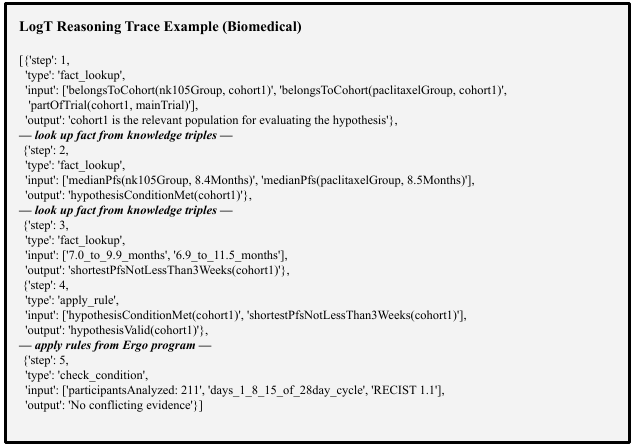}
\end{figure*}

\begin{figure*}[!t]
    \centering
    \includegraphics[width=\textwidth]{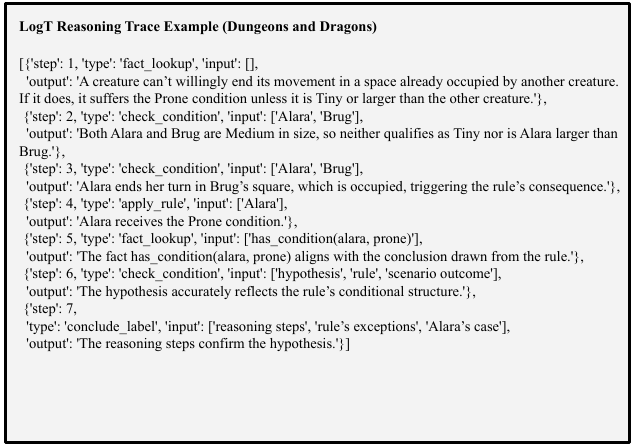}
\end{figure*}

\clearpage
\subsection{Appendix C.5 - Grounded LLM Evaluation}

\begin{figure}[h]
    \centering
    \includegraphics[width=\textwidth]{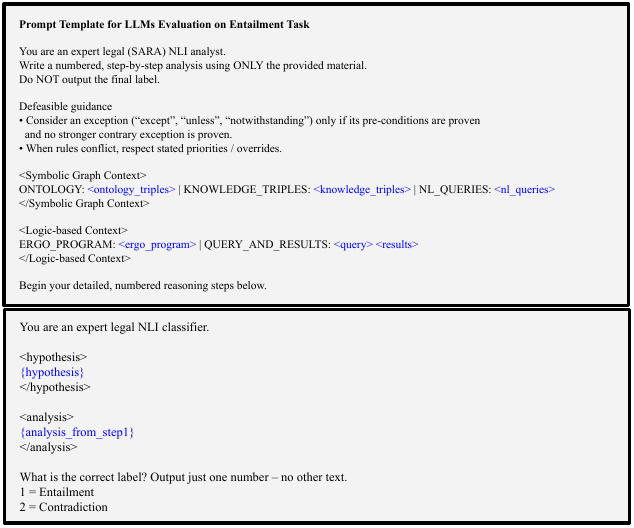}
    \label{fig:c5-ergo-gen}
\end{figure}

\clearpage
\section{Appendix D - Hypothesis Examples of Different Reasoning Mode}
\begin{figure}[h]
    \centering
    \includegraphics[width=1\textwidth]{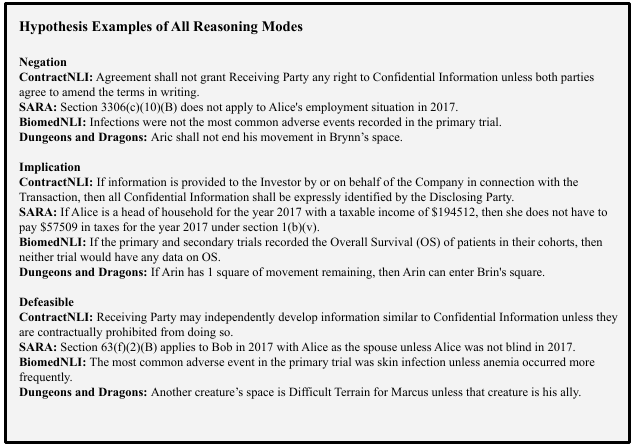}
    \caption{Hypothesis examples of all benchmarks and reasoning modes}
    \label{fig:d-hypothesis-examples}
\end{figure}

\clearpage

\section{Appendix E - ErgoAI Program Synthesis}
\label{app:ergo-syntax}

\begin{figure}[h]
    \centering
    \includegraphics[width=1\textwidth]{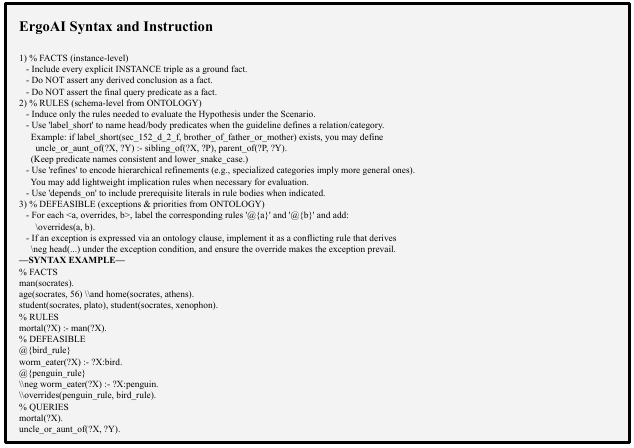}
    \caption{ErgoAI syntax and instruction that goes the ErgoAI program generation prompt.}
    \label{fig:ergo-instruction}
\end{figure}

To guide Ergo Program synthesis from knowledge triples and NL queries, we included valid ErgoAI syntax examples in the LLM prompt and specific instructions on how to convert to valid syntax. More specifically, as shown in Fig. \ref{fig:ergo-instruction}, we include ErgoAI syntax examples of facts, rules, defeasible rules, and queries. To parse facts into entries into an ErgoAI rule bank, we specifically instruct the model to distinguish between three main categories of logical constructs based on the input knowledge structure.

The mapping follows a structured approach where facts represent instance-level assertions extracted directly from explicit triples without deriving conclusions, rules encode schema-level relationships from the ontology using consistent predicate naming conventions and hierarchical refinements, and defeasible rules handle exceptions and priority relationships through labeled rule conflicts and override declarations. For instance, the bird-penguin example demonstrates this mechanism by establishing a general rule that birds eat worms (@\{bird\_rule\}) and an exception rule that penguins do not (@\{penguin\_rule\}), with the \textbackslash overrides(penguin\_rule, bird\_rule) declaration ensuring the exception takes precedence when both rules are applicable. 




\clearpage               
\section*{Appendix F – Benchmark Statistics}
\label{app:stats}

Detailed statistics for the four benchmarks used to evaluate \textsc{LogT} are presented in Table \ref{tab:benchmark-summary}. To ensure sufficient representation across the three reasoning modes (Negation, Implication, and Defeasibility), we curated the final test sets.

\begin{itemize}
    \item \textbf{BiomedNLI} evaluates reasoning in medical contexts by classifying statements as entailment or contradiction based on clinical trial reports. We retain the same number of the original test set of 200 instances.
    \item \textbf{ContractNLI} is a legal domain dataset containing document-level evidence within contract clauses, annotated with entailment, contradiction, and neutral labels. We sampled 1,000 diverse examples from the original test set.
    \item \textbf{SARA} consists of scenarios for statutory reasoning over the U.S. Internal Revenue Code, requiring application of complex legal rules. We retain the same number of the original test set of 261 instances.
    \item \textbf{D\&D} is our newly introduced benchmark based on the Dungeons \& Dragons rulebook, consisting of 149 hand-crafted examples reflecting logic-based gameplay reasoning.
\end{itemize}

\begin{table}[h!]
  \centering
  \small
  \renewcommand{\arraystretch}{1.3}
  \setlength{\tabcolsep}{6pt}
  \begin{tabular}{@{}lccrrr@{}}
    \toprule
    \textbf{Benchmark} & \textbf{Domain} & \textbf{Total} & \textbf{Neg.} & \textbf{Imp.} & \textbf{Deaf.} \\
    \midrule
    BiomedNLI   & Med.  & 200  & 84  & 67  & 49 \\
    ContractNLI & Legal & 1000 & 463 & 223 & 314 \\
    SARA        & Legal & 261  & 86  & 86  & 89 \\
    D\&D        & Games & 149  & 50  & 50  & 49 \\
    \bottomrule
  \end{tabular}
  \caption{Benchmark statistics by domain and reasoning type.}
  \label{tab:benchmark-summary}
\end{table}

\section{Appendix G - Pilot Study Results and Enhanced Benchmarks Validation Statistics}
\label{app:pilot-study}

To assess whether existing NLI benchmarks remain sufficiently challenging for modern LLMs, we conducted a pilot study using the original hypotheses from the \textbf{ContractNLI} and \textbf{BiomedNLI} datasets. Our analysis covers both findings from prior literature and results from our own baseline prompting experiments.

\paragraph{Findings from Existing Literature.}
Several studies have already shown that top-tier models perform remarkably well on these benchmarks. Table~\ref{tab:pilot-lit} summarizes key reported results.

\begin{table}[h]
\centering
\small
\renewcommand{\arraystretch}{1.2}
\begin{tabular}{@{}llp{5cm}@{}}
\toprule
\textbf{Dataset} & \textbf{Domain} & \textbf{Reported Performance} \\
\midrule
SARA         & Legal    & GPT-4 (0.868) \cite{blair2023can,holzenberger2023connecting} \\
ContractNLI  & Legal    & GPT-4 (0.82–0.96; varies by contract type) \cite{narendra2024enhancing,yao-etal-2024-samples} \\
BiomedNLI       & Medical  & LLaMA 70B (0.793) \\
\bottomrule
\end{tabular}
\caption{Performance of large models on original hypotheses from standard benchmarks (as reported in prior work).}
\label{tab:pilot-lit}
\end{table}

\paragraph{Simple Prompting Results.}
To validate these trends, we ran a basic prompting baseline on the original ContractNLI test set across a range of models. Results are shown in Table~\ref{tab:pilot-baseline}.

\begin{table}[h]
\centering
\small
\renewcommand{\arraystretch}{1.3}
\begin{tabular}{@{}lcc@{}}
\toprule
\textbf{Model} & \textbf{ContractNLI} & \textbf{BiomedNLI} \\
\midrule
LLaMA 3.3 70B   & 0.902 & 0.812 \\
GPT-4o          & 0.861 & 0.901 \\
GPT-4o mini     & 0.811 & 0.856 \\
\bottomrule
\end{tabular}
\caption{Simple prompting accuracy on original hypotheses from ContractNLI and BiomedNLI.}
\label{tab:pilot-baseline}
\end{table}

The results suggest that standard NLI benchmarks may no longer pose sufficient challenge for modern large language models. Even with minimal prompt engineering, models such as LLaMA 3.3 and GPT-4o achieve high accuracy, indicating that these datasets may not adequately differentiate models' reasoning capabilities. This highlights a saturation point in benchmark utility and underscores the need for more diagnostically precise evaluations. Motivated by this, we construct a set of enhanced benchmarks that explicitly probe reasoning mode such as negation, implication, and defeasible reasoning.

\paragraph{Validation with Standard NLI Models.}
To further verify the increased difficulty of our newly generated hypotheses, we evaluated both the original and enhanced hypotheses using two standard NLI models: RoBERTa-NLI and BERT-NLI. As shown in Table \ref{tab:nli-hardness}, both models performed worse on the enhanced hypotheses across all benchmarks.

\paragraph{Human Evaluation of Enhanced Benchmarks.}
To complement automated evaluation and ensure the validity of our enhancement, we sampled 100 examples from all enhanced benchmarks for manual quality check. These samples were assessed for semantic clarity, logical soundness, and consistency with the intended reasoning phenomena. A subset of these examples, particularly those targeting defeasible reasoning, implication, and negation. Illustrative examples from this evaluation process are shown in Figure \ref{fig:d-hypothesis-examples}.

\begin{table}[h]
\centering
\small
\renewcommand{\arraystretch}{1.3}
\begin{tabular}{@{}lccc@{}}
\toprule
\textbf{Model} & \textbf{ContractNLI} & \textbf{BioMedNLI} & \textbf{SARA} \\
\midrule
\multicolumn{4}{l}{\textit{Original Hypotheses}} \\
RoBERTa-NLI & 0.681 & 0.659 & 0.702 \\
BERT-NLI    & 0.642 & 0.633 & 0.671 \\
\addlinespace[4pt]
\multicolumn{4}{l}{\textit{Enhanced Hypotheses}} \\
RoBERTa-NLI & 0.512 & 0.486 & 0.529 \\
BERT-NLI    & 0.478 & 0.441 & 0.501 \\
\bottomrule
\end{tabular}
\caption{Accuracy of off-the-shelf NLI models on original vs. enhanced hypotheses across three benchmarks.}
\label{tab:nli-hardness}
\end{table}

\clearpage

    



    

\section{Appendix H - Additional Evaluation Results}

\subsection{Appendix H.1 -  LOGT Performance against the
Average Performance of All Baselines}

\begin{figure}[htbp]
    \centering
    \includegraphics[width=\columnwidth]{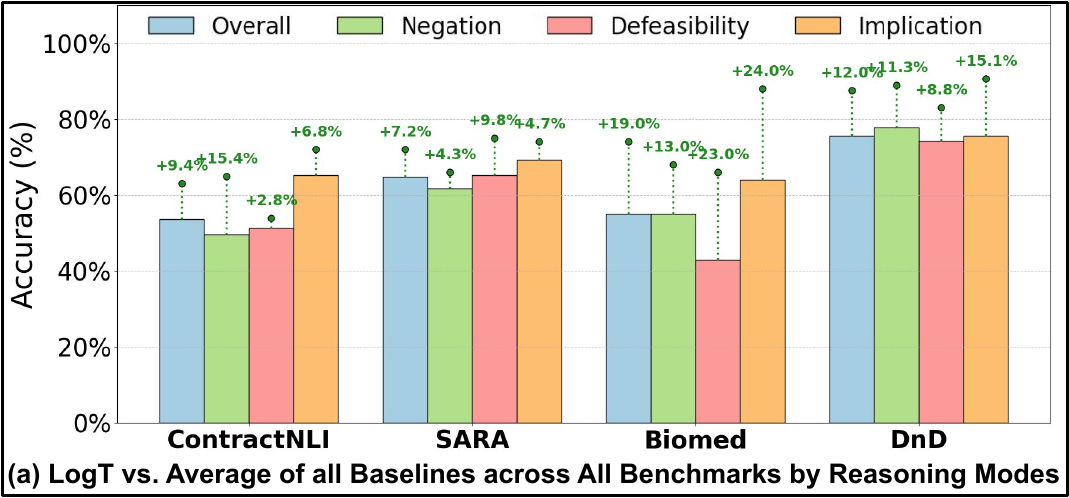}
    \caption{LogT+reasoning accuracy against baseline reasoning accuracy.}
    \label{fig:h1-result-figure-appendix}
\end{figure}

Based on the results shown in Fig.~\ref{fig:h1-result-figure-appendix}, when comparing LogT against the average performance of all baselines rather than the strongest individual baseline, our approach demonstrates consistent and substantial improvements across all benchmarks and reasoning modes. LogT achieves notable gains ranging from +2.8\% to +24.0\% across different datasets. The improvements are most pronounced on the Biomed dataset, where LogT shows exceptional gains of +24.0\% in negation and +23.0\% in implication reasoning modes. These results highlight LogT's robustness across diverse reasoning tasks. The consistent improvements in all four reasoning modes demonstrate that LogT's structured knowledge synthesis approach provides systematic advantages over existing methods.

\subsection{Appendix H.2 -  Reasoning Trace Instance-level Analysis}

\begin{figure}[htbp]
    \centering
    \begin{subfigure}[t]{0.45\textwidth}
        \includegraphics[width=\linewidth]{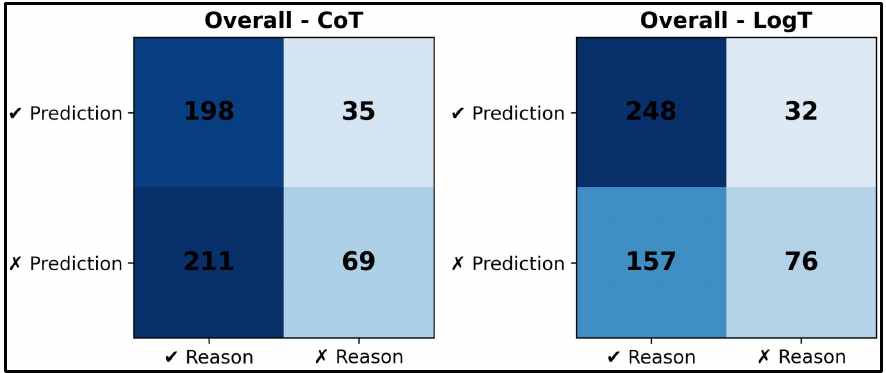}
        \caption{Overall}
    \end{subfigure}

    \begin{subfigure}[t]{0.45\textwidth}
        \includegraphics[width=\linewidth]{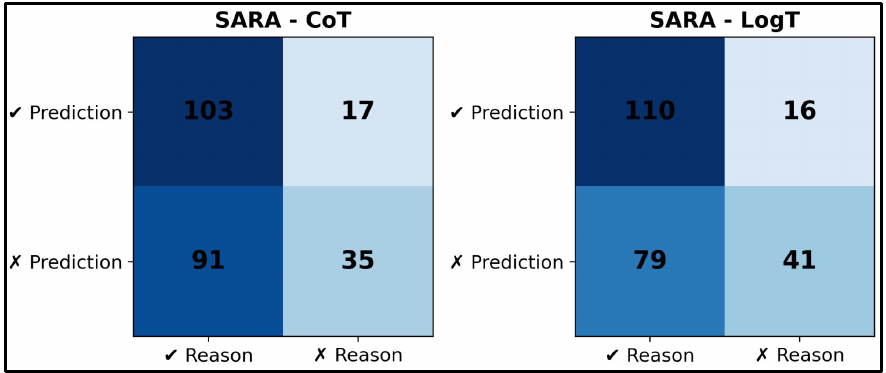}
        \caption{SARA}
    \end{subfigure}
    
    \vspace{0.3em}
    
    \begin{subfigure}[t]{0.45\textwidth}
        \includegraphics[width=\linewidth]{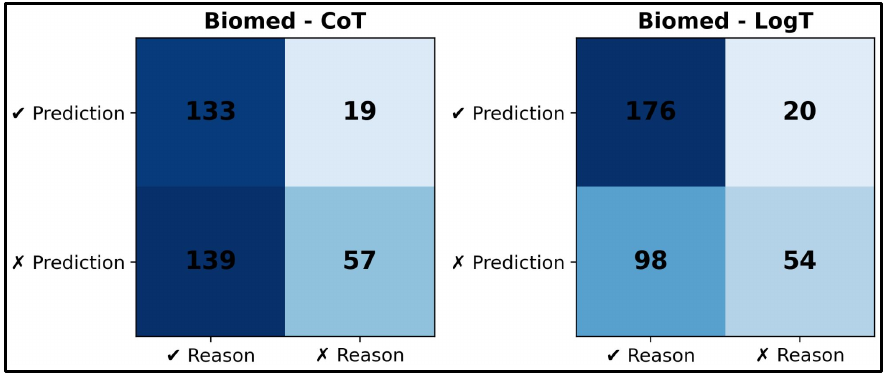}
        \caption{Biomed}
    \end{subfigure}
    \hfill
    \begin{subfigure}[t]{0.45\textwidth}
        \includegraphics[width=\linewidth]{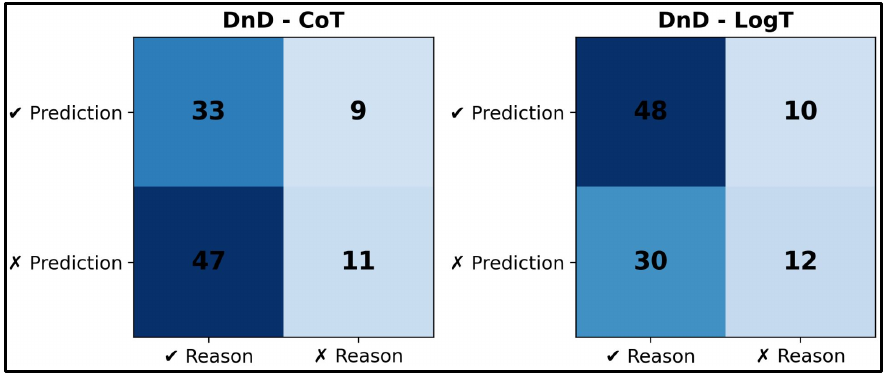}
        \caption{D\&D}
    \end{subfigure}
    
    \caption{Comparison of confusion matrices for CoT and LogT across three benchmarks; (a) Overall, (b) SARA, (c) BiomedNLI, and (d) D\&D. \textsc{LogT} improves the alignment between correct predictions and correct reasoning traces.}
    \label{fig:conf-all}
\end{figure}

We analyze the reasoning traces of LogT in comparison to baseline CoT reasoning and find improved alignment between the correct predictions produced by LogT with correct reasoning traces.
Using our \textit{LLM-as-a-judge} setup to categorize reasoning traces and assign correct or incorrect reasoning labels, we find that LogT has correct reasoning traces for 89\% of its correct predictions, improving on CoT's 85\% alignment overall (see Fig. \ref{fig:conf-all}).
We observe similar improvements when we break this down by dataset.
86\% alignment for correct prediction to 87\% for SARA, 88\% to 90\% for Biomed, and 79\% to 83\% for D\&D.

\begin{figure}[htbp]
    \centering
    \begin{subfigure}[t]{0.45\textwidth}
        \includegraphics[width=\linewidth]{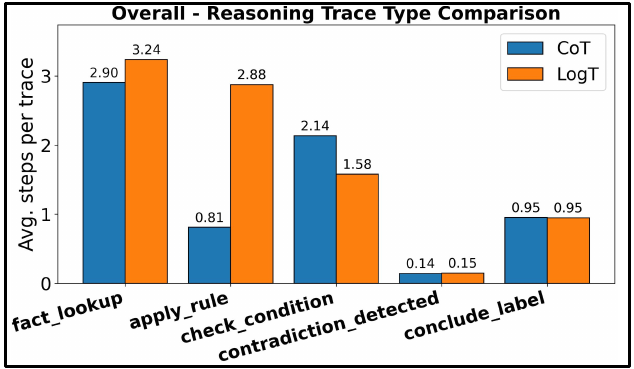}
        \caption{Overall}
    \end{subfigure}
    \hfill
    \begin{subfigure}[t]{0.45\textwidth}
        \includegraphics[width=\linewidth]{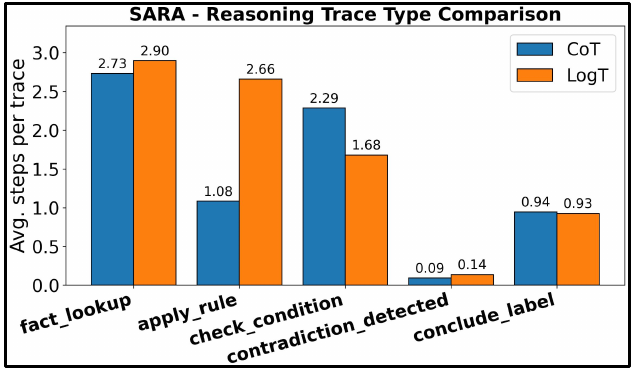}
        \caption{SARA}
    \end{subfigure}
    
    \vspace{0.3em}
    
    \begin{subfigure}[t]{0.45\textwidth}
        \includegraphics[width=\linewidth]{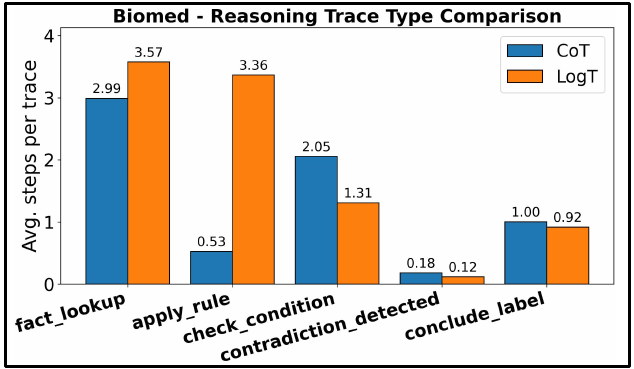}
        \caption{Biomed}
    \end{subfigure}
    \hfill
    \begin{subfigure}[t]{0.45\textwidth}
        \includegraphics[width=\linewidth]{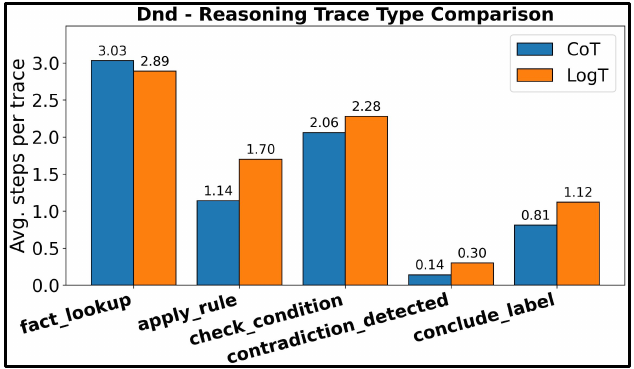}
        \caption{DnD}
    \end{subfigure}
    
    \caption{Comparison of reasoning trace types and their average steps per trace from our approach LogT and baseline CoT across datasets, based on \textit{LLM-as-a-judge} annotations. We observe that LogT tends to use more \textit{apply\_rule} steps irrespective of dataset. \textit{check\_condition} steps produced by LogT are shorter for most datasets, except for D\&D, where LogT produces slightly more such steps than CoT. Likewise, \textit{fact\_lookup} steps are longer for LogT for most datasets, except for D\&D.}
    \label{fig:trace-all}
\end{figure}

We further break down the reasoning traces to look for the qualitative differences between our LogT approach and CoT (see Fig. \ref{fig:trace-all}).
We find that the greatest difference between the reasoning traces of the two methods is a greater emphasis placed on \textit{apply\_rule} steps, which are given 2.88 steps on average by the LogT predictions compared to just 0.81 in the CoT traces.
This trend holds when we breakdown the analysis down into individual datasets and suggests that contexts provided from LogT forces models to adhere to the given ontology.

Interestingly, the other big differences between the traces, more steps devoted to \textit{fact\_lookup} (with 2.9 steps on average from the CoT approach and 3.24 steps on average from LogT overall) and fewer steps devoted to \textit{check\_condition} reasoning (2.14 steps on average for CoT and just 1.58 for LogT) by LogT do not hold for all of our individual datasets.
We still observe more \textit{fact\_lookup} steps and fewer \textit{check\_condition} steps in the SARA and Biomed datasets, but the opposite holds for the D\&D dataset.
This points to value of our custom D\&D dataset, which may require different reasoning for LLMs to successfully solve it.

\subsection{Appendix H.3 - Reasoning Trace Qualitative Analysis}
\label{app:trace-analysis}

To understand the qualitative differences between Chain-of-Thought (CoT) prompting and our proposed \textsc{LogT}, we compare their generated reasoning traces on a set of representative problems. Our analysis focuses on how each model structures its thought process, decomposes complex reasoning steps, and ultimately arrives at a solution. In these examples, \textsc{LogT} receives additional structured context—either symbolic or logical—while CoT relies solely on natural language reasoning. We highlight key differences in how models approach problem decomposition, symbol grounding, and final answer extraction, particularly in settings that require multi-step inference or domain-specific knowledge.

Across both panels in Table~\ref{tab:reasoning-trace-examples}, we observe that \textsc{LogT} consistently produces more structured, faithful reasoning traces compared to Chain-of-Thought (CoT) prompting. While CoT often approximates logical inferences through informal natural language steps, it tends to leave out critical rule applications or misinterpret negations, as seen in examples involving marital status under §7703(b) or eligibility based on HER2 expression. In contrast, \textsc{LogT} explicitly grounds each step in retrieved facts (e.g., marriedTo(alice, bob) or hasGivenSomethingForServices(alice, bob)) and applies formal inference rules, resulting in accurate conclusions with transparent justification. The rightmost columns of both panels further confirm that CoT frequently fails to predict the correct label, while \textsc{LogT} succeeds across all examples.

The examples in the second panel of Table~\ref{tab:reasoning-trace-examples-2} further highlight \textsc{LogT}’s superiority in handling negation, conditional logic, and statute interpretation. In the Section 2(b)(2)(C) example, CoT provides a vague eligibility judgment, whereas \textsc{LogT} correctly falsifies the hypothesis by checking for remarriage and applying conjunctive rule conditions. Similarly, in the Section 3306(b)(7) case, CoT treats the exchange of services imprecisely, while \textsc{LogT} uses symbolic inputs to formalize the condition and apply the correct logical negation. These examples underscore the importance of integrating structured symbolic reasoning into language model pipelines, particularly for domains requiring legal precision, multi-hop deduction, or verifiable justification.

\begin{table*}[t]
\small
\renewcommand{\arraystretch}{1.35}
\setlength{\tabcolsep}{4pt}
\centering
\begin{tabularx}{\textwidth}{@{}l X X c c@{}}
\toprule
\textbf{Model} & \textbf{CoT Trace} & \textbf{LogT Trace} & \textbf{CoT Prediction} & \textbf{LogT Prediction} \\
\midrule

Claude-Haiku &
\textbf{Step 1:} Identify Key Conditions of §7703(b) \newline
\textbf{Step 2:} Alice married Bob in 2018, filed separately $\rightarrow$ meets first two conditions \newline
\textbf{Step 3:} Maintains home for Charlie, paid over half costs \newline
\textbf{Step 4:} Bob lived at home from Sept 2017 to Feb 2018 \newline
\textbf{Step 5:} Bob was member > 6 months $\rightarrow$ fails condition \newline
\textbf{Step 6:} Conclude: §7703(b)(3) does NOT apply to Alice &
\textbf{Step 1:} Formalize: married + sep return + spouse not present in last 6 months \newline
\textbf{Step 2:} \texttt{marriedTo(alice, bob)} \newline
\textbf{Step 3:} \texttt{maintainsHomeFor(alice, charlie)}, \texttt{entitledToDeductionFor} \newline
\textbf{Step 4:} \texttt{notMemberOfHousehold(bob, last\_6\_months\_of\_2018)} \newline
\textbf{Step 5:} Rule fires $\rightarrow$ sectionApplies(section\_7703b3, alice) \newline
\textbf{Step 6:} Conclude: §7703(b)(3) likely applies to Alice &
\ding{55} & \ding{51} \\

Llama-3.3-70B &
\textbf{Step 1:} Alice married Bob (April 5, 2012) $\rightarrow$ considered married for 2018 \newline
\textbf{Step 2:} Files separate return \newline
\textbf{Step 3:} Maintains home for Charlie since Sept 16, 2017 \newline
\textbf{Step 4:} Furnished over half the cost (assumed) \newline
\textbf{Step 5:} Bob not in household from Feb 10 – Dec 31, 2018 \newline
\textbf{Step 6:} Says 3 of 4 conditions met; missing explicit confirmation of cost \newline
\textbf{Step 7:} Concludes: §7703(b) applies likely but uncertain &
\textbf{Step 1:} \texttt{maintainsHomeForYear(alice, 2018)} via home > half-year \newline
\textbf{Step 2:} \texttt{marriedTo(alice, bob)} \newline
\textbf{Step 3:} \texttt{maintainsHomeFor(alice, charlie)}, \texttt{entitledToDeductionFor(alice, charlie)} \newline
\textbf{Step 4:} \texttt{spouseNotMemberOfHousehold(alice, last\_6\_months\_of\_2018)} \newline
\textbf{Step 5:} Fires rule: sectionApplies(section\_7703b3, alice) \newline
\textbf{Step 6:} Conclusion: §7703(b)(3) applies to Alice \newline
\textbf{Step 7:} Final step confirms hypothesis is supported &
\ding{55} & \ding{51} \\

Deepseek-R1 &
\textbf{Step 1:} Lookup HER2- breast cancer and ErbB2(HER2) overexpression \newline
\textbf{Step 2:} HER2 overexpression is a hallmark of HER2+ cancer \newline
\textbf{Step 3:} HER2- status means no HER2 overexpression \newline
\textbf{Step 4:} Trial requires HER2 overexpression \newline
\textbf{Step 5:} Apply logic: HER2- cancer does not satisfy HER2+ criterion \newline
\textbf{Step 6:} Conclude: HER2- and HER2+ are mutually exclusive $\rightarrow$ contradiction with eligibility &
\textbf{Step 1:} Check condition mismatch: HER2- vs HER2+ eligibility \newline
\textbf{Step 2:} Lookup: overexpression is required for trial eligibility \newline
\textbf{Step 3:} HER2- patients would not meet inclusion criteria \newline
\textbf{Step 4:} Apply contradiction rule: HER2- patient with HER2+ trial \newline
\textbf{Step 5:} Conflict with trial requirements confirmed \newline
\textbf{Step 6:} Identify mutually exclusive biological states \newline
\textbf{Step 7:} Conclude: aligns with hypothesis that trial requirements not met &
\ding{55} & \ding{51} \\

Deepseek-R1 &
\textbf{Step 1–2:} Identify dosing regimen for Letrozole in primary trial \newline
\textbf{Step 3:} Confirm sentinel lymph node biopsy in secondary trial \newline
\textbf{Step 4–5:} Eligibility rules link interventions to trials \newline
\textbf{Step 6–8:} Apply to patients 1–3, showing treatment mappings \newline
\textbf{Step 9–10:} Verify protocol adherence per trial \newline
\textbf{Step 11–12:} Each hypothesis clause aligns with relevant info \newline
\textbf{Step 13:} Final check confirms no contradictions &
\textbf{Step 1–2:} Identify Letrozole and biopsy interventions \newline
\textbf{Step 3:} Link eligibility to trial interventions \newline
\textbf{Step 4:} No explicit eligibility criteria found \newline
\textbf{Step 5:} Secondary trial intervention lacks specificity \newline
\textbf{Step 6:} Primary trial aligns with hypothesis \newline
\textbf{Step 7:} Final output: hypothesis cannot be fully validated due to missing details &
\ding{55} & \ding{51} \\

\bottomrule
\end{tabularx}
\caption{Qualitative comparison of reasoning traces produced by Chain-of-Thought (CoT) and \textsc{LogT} for statutory tax reasoning tasks. CoT fails to satisfy or confirm all required conditions in both cases, while \textsc{LogT} correctly applies formal rules and reaches the correct conclusion.}
\label{tab:reasoning-trace-examples}
\end{table*}

\begin{table*}[t]
\small
\renewcommand{\arraystretch}{1.35}
\setlength{\tabcolsep}{4pt}
\centering
\begin{tabularx}{\textwidth}{@{}l X X c c@{}}
\toprule
\textbf{Model} & \textbf{CoT Trace} & \textbf{LogT Trace} & \textbf{CoT Prediction} & \textbf{LogT Prediction} \\
\midrule

GPT-4o-mini &
\textbf{Step 1–3:} Establish family relationships (Charlie is Alice’s father, Bob is Charlie’s brother) \newline
\textbf{Step 4–5:} Apply rules to derive that Alice is a descendant of Bob’s sibling \newline
\textbf{Step 6:} Check the negation condition and falsify it \newline
\textbf{Step 7:} Conclude the hypothesis is supported &
\textbf{Step 1:} Define section 152(d)(2)(E) \newline
\textbf{Step 2–3:} Map family structure (Bob–Charlie–Alice) \newline
\textbf{Step 4:} Misread or test a negated condition \newline
\textbf{Step 5–6:} Reasoning becomes unclear—confuse relationship alignment \newline
\textbf{Step 7:} Conclude ambiguity remains in whether Alice qualifies &
\ding{55} & \ding{51} \\

LLaMA-3.3-70B &
\textbf{Step 1:} Read section 2(b)(2)(C) $\rightarrow$ conditions for surviving spouse \newline
\textbf{Step 2:} Bob has not remarried after Alice's death \newline
\textbf{Step 3:} Assumed residents, joint return applies \newline
\textbf{Step 4:} Bob qualifies as surviving spouse in 2014 \newline
\textbf{Step 5:} Limitation clause likely not violated \newline
\textbf{Step 6:} Hypothesis judged plausible \newline
\textbf{Step 7:} Conclusion: hypothesis is supported &
\textbf{Step 1:} \texttt{diedOn(alice, july\_9\_2014)} and \texttt{marriedTo(alice, bob)} \newline
\textbf{Step 2:} -> \texttt{spouseDiedDuringTaxableYear(bob, 2014)} \newline
\textbf{Step 3:} Checks that Bob was not remarried \newline
\textbf{Step 4:} Fires rule: \texttt{jointReturnPossible(bob, 2014)} \newline
\textbf{Step 5:} All rule inputs met -> \texttt{section\_2b2c\_applies(bob, 2014)} \newline
\textbf{Step 6:} Conclusion: §2(b)(2)(C) applies to Bob \newline
\textbf{Step 7:} Contradiction check falsified negation &
\ding{55} & \ding{51} \\

Claude-Haiku &
\textbf{Step 1:} Defines 'employer' and wage conditions \newline
\textbf{Step 2:} Alice gave Bob a typewriter valued at \$323; Bob painted her house \newline
\textbf{Step 3:} Interprets exchange as non-cash compensation \newline
\textbf{Step 4:} Recognizes exchange occurred outside workplace duties \newline
\textbf{Step 5:} Suggests statutory implications may arise \newline
\textbf{Step 6:} Concludes that further analysis of Section 3306(b)(7) is needed &
\textbf{Step 1:} \texttt{hasGivenSomethingForServices(alice, bob)} established \newline
\textbf{Step 2:} \texttt{Section 3306(b)(7)} retrieved – wage definitions \newline
\textbf{Step 3:} Applies special rule (paragraph 4) to non-standard wage \newline
\textbf{Step 4:} Confirms Alice gave a typewriter in exchange for services \newline
\textbf{Step 5:} Condition: Section applies unless Alice \emph{never} gave anything \newline
\textbf{Step 6:} Applies negation logic: Alice has given something \newline
\textbf{Step 7:} Concludes Section 3306(b)(7) likely applies &
\ding{55} & \ding{51} \\

\bottomrule
\end{tabularx}
\caption{Qualitative comparison of reasoning traces produced by Chain-of-Thought (CoT) and \textsc{LogT} for statutory tax reasoning tasks. CoT fails to satisfy or confirm all required conditions in both cases, while \textsc{LogT} correctly applies formal rules and reaches the correct conclusion.}
\label{tab:reasoning-trace-examples-2}
\end{table*}
\end{document}